\documentclass{article}
\usepackage{spconf,amsmath,amssymb,amsfonts,graphicx}
\usepackage[export]{adjustbox}
\usepackage{subfig}
\usepackage{algorithmic}
\graphicspath{{figs/}}
\usepackage{textcomp}
\usepackage{xcolor}
\usepackage{cite}
\usepackage{siunitx}
\usepackage{booktabs}
\usepackage{multirow}
\usepackage{mathtools}
\usepackage{rotating}
\usepackage{array}

\usepackage{xspace}
\makeatletter
\DeclareRobustCommand\onedot{\futurelet\@let@token\@onedot}
\def\@onedot{\ifx\@let@token.\else.\null\fi\xspace}

\def\ie{{\it i.e}\onedot}

\makeatother

\title{Text-Guided Scene Sketch-to-Photo Synthesis}
\name{\begin{tabular}{c}AprilPyone MaungMaung$^{\star}$ \quad Makoto Shing$^{\star}$ \quad Kentaro Mitsui$^{\star}$ \quad Kei Sawada$^{\star}$ \quad
Fumio Okura$^{\dagger}$\end{tabular}}

\address{$^{\star}$ rinna Co., Ltd. \qquad
    $^{\dagger}$Osaka University}

\makeatletter

\begin{document}
%
\maketitle
\begin{abstract}
We propose a method for scene-level sketch-to-photo synthesis with text guidance. Although object-level sketch-to-photo synthesis has been widely studied, whole-scene synthesis is still challenging without reference photos that adequately reflect the target style.
To this end, we leverage knowledge from recent large-scale pre-trained generative models, resulting in text-guided sketch-to-photo synthesis without the need for reference images.  
To train our model, we use self-supervised learning from a set of photographs. Specifically, we use a pre-trained edge detector that maps both color and sketch images into a standardized edge domain, which reduces the gap between photograph-based edge images (during training) and hand-drawn sketch images (during inference).
We implement our method by fine-tuning a latent diffusion model (\ie, Stable Diffusion) with sketch and text conditions. Experiments show that the proposed method translates original sketch images that are not extracted from color images into photos with compelling visual quality.
\end{abstract}
\section{Introduction}
\label{sec:intro}
We experience sketching at least once in a lifetime from childhood to professionals.
It is desirable to have a model that automatically generates realistic diverse images from human-drawn (freehand) sketches.
With the recent development of generative models, there have been many works that attempted to generate photorealistic images from sketches~\cite{liu2020unsupervised,richardson2021encoding,xiang2022adversarial,isola2017image}.
However, these works mainly focus on categorical object-level sketches, thus generating photorealistic images from scene-level sketches is still challenging.
As mentioned in~\cite{wang2022unsupervised}, key issues of scene-level sketch translation are (1) lack of training data and (2) complexity of scene-level sketches.
A potential solution for scene-level synthesis is to build a composite dataset in which isolated objects are composed to make up a scene~\cite{gao2020sketchycoco}.
However, this approach is hard to generalize because detecting objects from sketches is difficult, and a scene may contain out-of-categories objects~\cite{wang2022unsupervised}.

To overcome the data shortage and scene complexity in scene-level sketch-to-photo (S2P) synthesis, we propose to leverage the knowledge of large pre-trained generative models, such as Stable Diffusion~\cite{rombach2022high}.
Unlike recent S2P works using reference photos~\cite{wang2022unsupervised,liu2021self}, we use text guidance that provides users an easier alternative to control the style of the output image.
Specifically, we fine-tune off-the-shelf Stable Diffusion models by conditioning with sketches and text descriptions, inspired by a depth-to-image diffusion model~\cite{rombach2022high} that translates depth maps to images with text guidance.
Since paired sketches and color images are not available, we adopt the use of a standardized domain as in~\cite{wang2022unsupervised} by utilizing a pre-trained edge detector as a pre-processing.

Experiments show that our fine-tuned model achieved lower Perceptual Image Patch Similarity (LPIPS)~\cite{zhang2018unreasonable} scores (\ie, higher similarity) compared to a recent diffusion-based S2P model~\cite{wang2022pretraining}.
In addition, our model also translates original sketch (line) images that are extracted from color images to realistic images with compelling visual quality.
Since, we leverage pre-trained large-scale text-to-image models, our model can control the style of the output with given text prompts naturally.

\section{Related Work}
\subsection{Sketch-to-Photo (S2P) Synthesis}

There are numerous works that attempt to generate photorealistic images from sketches~\cite{chen2018sketchygan,liu2020unsupervised,xiang2022adversarial,ghosh2019interactive}, in which, most studies focus on categorical object-level sketches.
For scene-level synthesis, SketchyScene~\cite{zou2018sketchyscene} proposed the first large-scale dataset of scene sketches by applying object segmentation and manually compositing scene sketches with reference to a cartoon image.
However, a recent work~\cite{wang2022unsupervised} pointed out that there is still a large domain gap to real scene sketches with reference to a real scene.
Therefore, current state-of-the-art scene S2P approaches utilize a reference image to generate photorealistic images~\cite{wang2022unsupervised,liu2021self}.

As diffusion models have dominated image synthesis, recent works~\cite{voynov2022sketch,cheng2023adaptively} achieve sketch-conditioned diffusion models via sketch-guided sampling with an additional guided model~\cite{voynov2022sketch} or with iterative latent variable refinement in pixel space~\cite{cheng2023adaptively}.
Unlike these recent S2P models, our method does not need any reference image and performs full sampling from pure noise. Moreover, our method does not have any additional guidance during sampling unlike previous diffusion-based methods~\cite{voynov2022sketch,cheng2023adaptively}.

As reference-free methods using sketch-conditioned diffusion models, PITI~\cite{wang2022pretraining} is the closest to ours. 
Our method differs from PITI in important ways: 1) PITI does not use text conditions, 2) PITI does not consider a standardized sketch domain but instead uses HED~\cite{xie2015holistically} sketches, and 3) PITI relies on the pre-trained two-stage diffusion models, namely a base model at the resolution of $64 \times 64$ and an upsampling model at the resolution of $256\times256$ (\ie, GLIDE~\cite{nichol2021glide}).
In contrast, our method utilizes a large-scale text-to-image model (\ie, Stable Diffusion~\cite{rombach2022high}) and naturally incorporates both sketch and text conditions.

As an unpublished concurrent study, a general image-to-image translation model achieves S2P tasks~\cite{control2023}. Although the use of large-scale pre-trained models is similar to our approach, we rather focus on a specific use case of the S2P task with human-drawn scene sketches, where we use domain standardization for effective self-supervised learning.

\subsection{Latent Diffusion Models}
Denoising diffusion probabilistic models (DDPM) have shown state-of-the-art results in image generation~\cite{dhariwal2021diffusion}.
Despite the great performance of diffusion models in generative modeling, diffusion models in pixel space consume an extensive amount of computational power.
Therefore, latent diffusion models (LDM) were proposed that worked in the latent space of a pre-trained variational autoencoder~\cite{rombach2022high}.
The success of LDM can be seen in the current trending text-to-image model, known as Stable Diffusion.
During sampling, classifier-free guidance~\cite{ho2022classifier} is often used to trade off between sample quality and diversity.
The output of the model with classifier-free guidance is
\begin{equation}
  \hat{\epsilon}_\theta(z_t|c) = \epsilon_\theta(z_t|\emptyset) + s \cdot (\epsilon_\theta(z_t|c) - \epsilon_\theta(z_t|\emptyset)), \label{eq:cfg}
\end{equation}
where $\epsilon_\theta$ is a learned network that predicts the noise, $z_t$ is a noisy latent, $c$ is a given condition, $s > 1$ is a guidance scale and $\emptyset$ is a null condition.
In this paper, we also deploy classifier-free guidance for sketch-to-photo synthesis.

\section{Methods}
Our text-guided S2P method leverages the diffusion models pre-trained with large image-text pairs.
In particular, inspired by the recent success of image-to-image tasks~\cite{saharia2022palette} using latent diffusion~\cite{rombach2022high} models, we are motivated to explore fine-tuning the Stable Diffusion\footnote{https://github.com/Stability-AI/stablediffusion} with scene sketch conditioning.

\begin{figure}[t]
\centering
\subfloat[Training.]{\includegraphics[width=\linewidth]{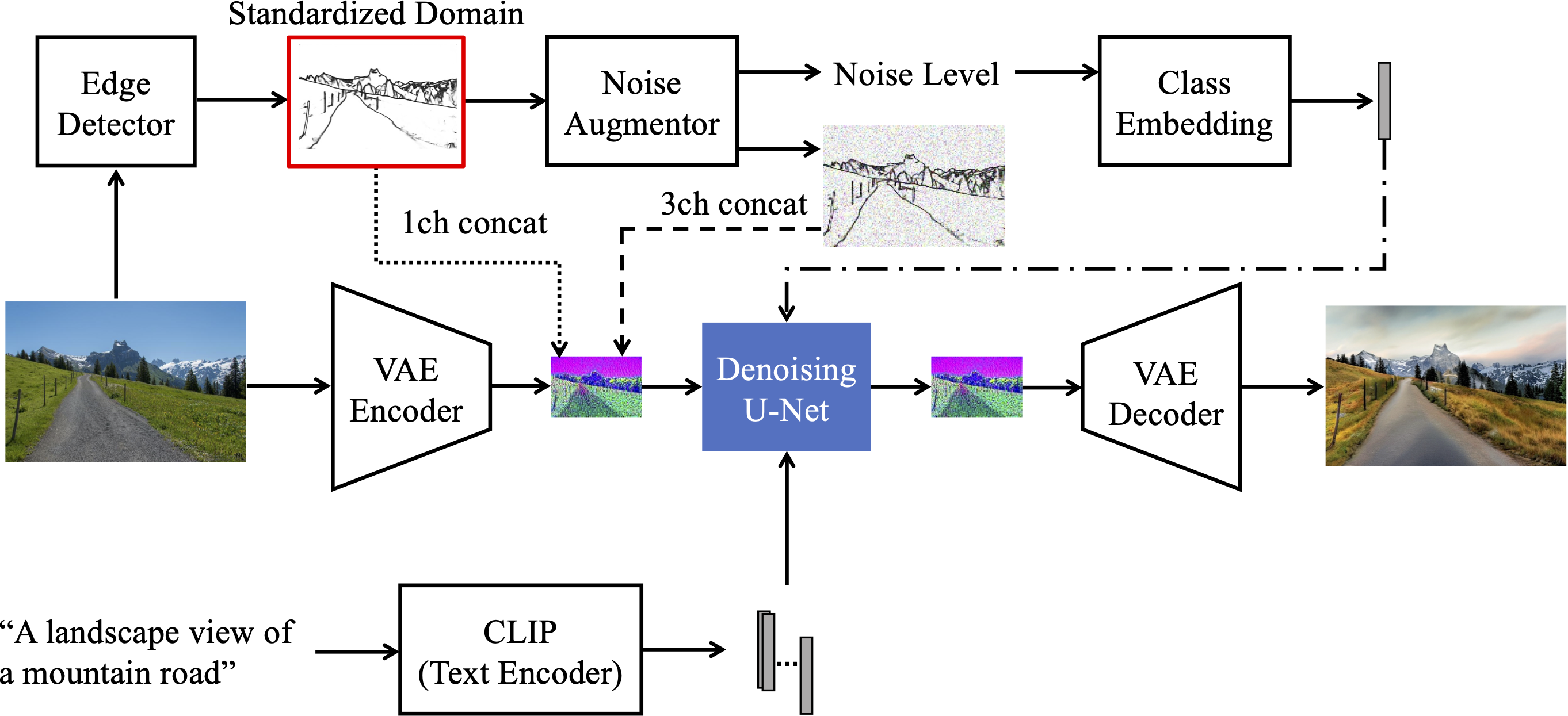}%
\label{fig:training}}
\hfil
\subfloat[Sampling.]{\includegraphics[width=\linewidth]{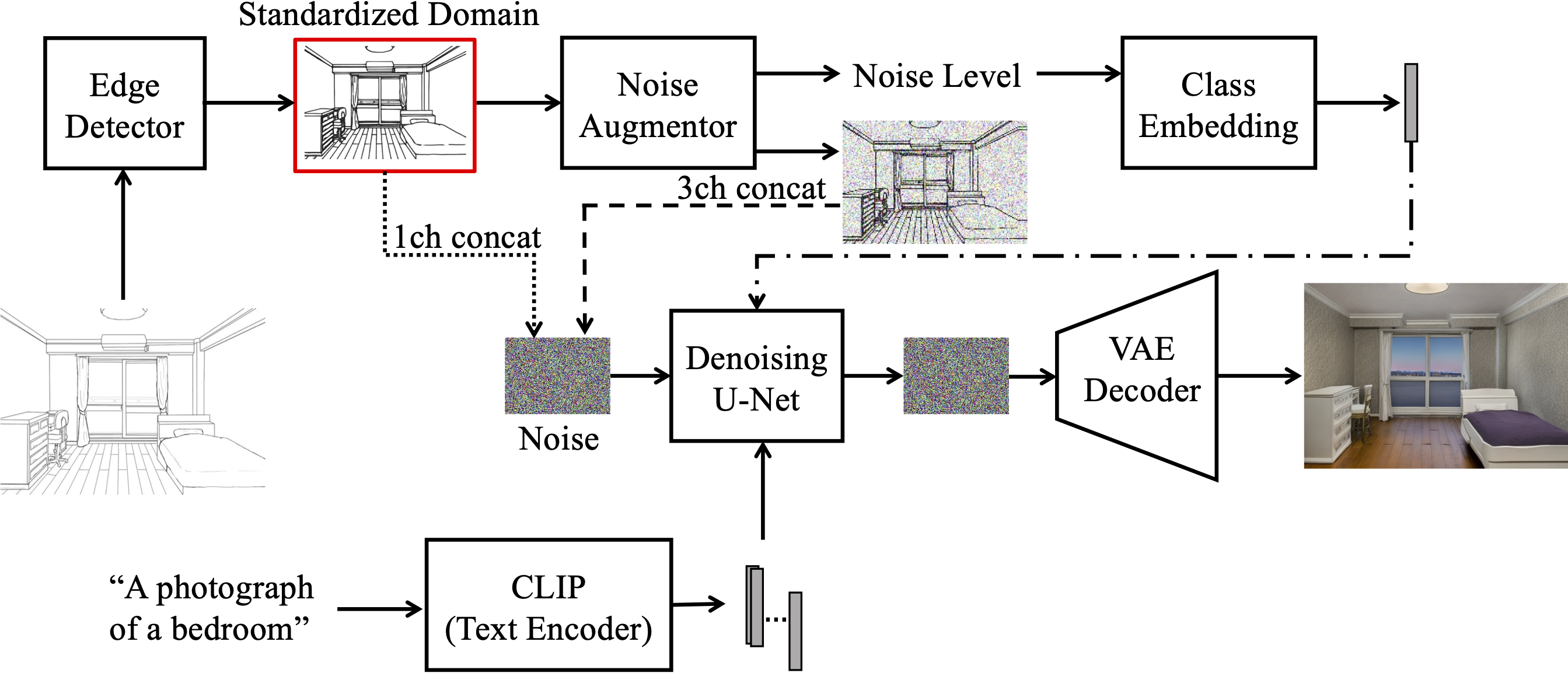}%
\label{fig:sampling}}
\caption{Overview of training and sampling of sketch conditioning on the Stable Diffusion. During training (a), only the blue highlighted block is updated and the rest of the blocks are fixed.\label{fig:overview}}
\end{figure}

An overview of sketch conditioning on the Stable Diffusion is depicted in Fig.~\ref{fig:overview}.
Given an input sketch image, we employ classifier-free guidance on the denoising process in the Stable Diffusion models, by concatenating the sketches to the input noise.
We achieve self-supervised learning from photographs (\ie, without hand-drawn sketches) by converting input images, whether photographs or sketches, into a standardized edge domain, inspired by a domain standardization technique described in~\cite{wang2022unsupervised}.

\subsection{Sketch Conditioning on Stable Diffusion}
We implement the sketch-conditioned diffusion models inspired by two implementations based on Stable Diffusion: Namely, one-channel concatenation similar to the depth-to-image translation, and three-channel concatenation similar to an upscaling model. Our experiments compare two models.

\vspace{2mm}\noindent{\bf One-channel concatenation:}
A depth-to-image translation model based on the Stable Diffusion 2.0 is released from Stability AI\footnote{https://huggingface.co/stabilityai/stable-diffusion-2-depth}. This model conditions the denoising process with monocular depth inferred by MiDaS~\cite{ranftl2020towards}.
As this model utilizes one-channel depth map concatenation to input as conditional information (see Fig.~\ref{fig:overview}), we design to replace depth maps with standardized sketches. Then, we fine-tune the model using our dataset to produce a sketch-to-photo model.

\vspace{2mm}\noindent{\bf Three-channel concatenation:}
Stability AI also released an x4 text-guided upscaling model\footnote{https://huggingface.co/stabilityai/stable-diffusion-x4-upscaler}.
In this model, three-channel low-resolution images are augmented with noises, then concatenated to input as conditional information.
In addition, noise level embedding is also provided as conditional information for this upscale model (see Fig.~\ref{fig:overview}).
We replace low-resolution images with standardized sketches and fine-tune the model for the sketch-to-photo task.

\subsection{Self-supervised Learning with Domain Standardization}
A nontrivial challenge of S2P synthesis is the lack of scene-level sketch-photo datasets.
It is thus intractable to learn a supervised model that generates photos from sketches.
Following the work by~\cite{wang2022unsupervised}, we employ a domain standardization technique using a pre-trained edge detector~\cite{poma2020dense} so that our model can be self-supervised just from a set of photographs.

\begin{figure}[!t]
\centering
\includegraphics[width=\linewidth]{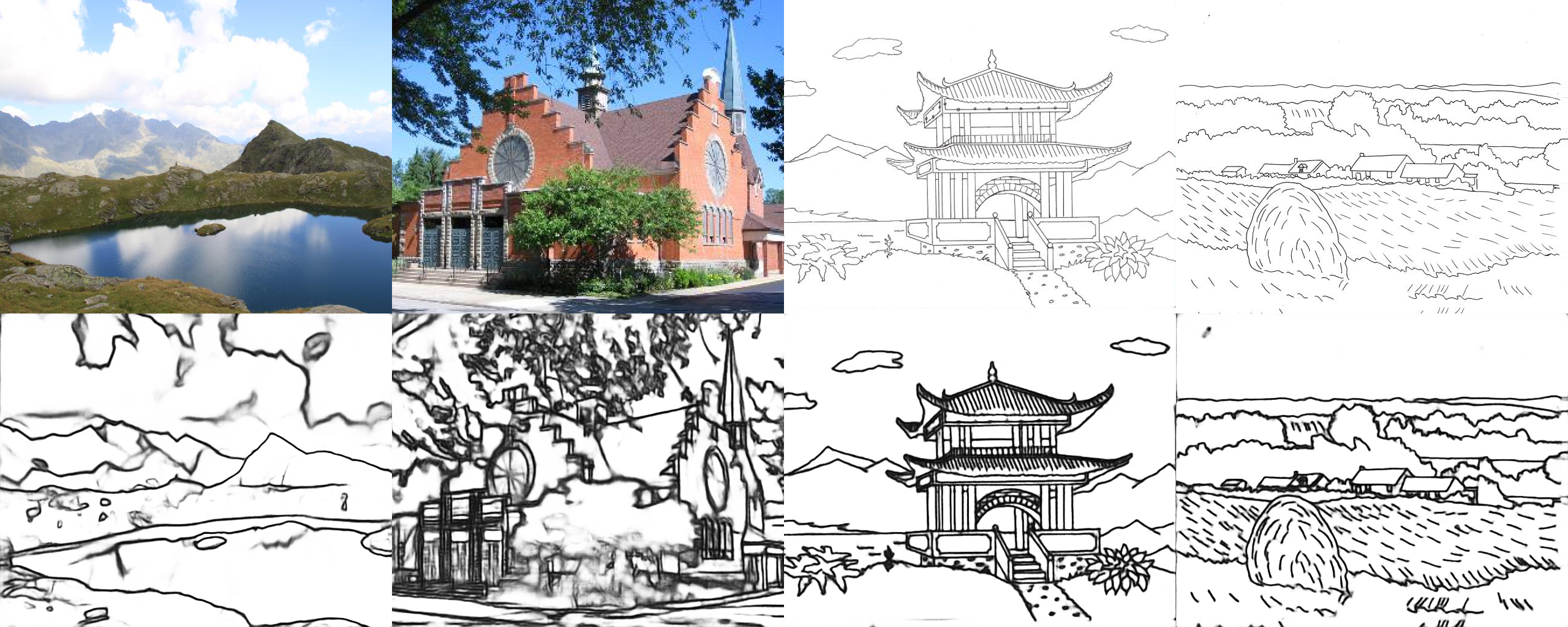}
\caption{Examples of standardized images. Both photos and sketches are converted to a standardized domain (second row) to narrow down the domain gap.\label{fig:edge}}
\end{figure}

The edge detector is trained to map both color photographs and sketches to a common domain (\ie, edge maps).
The use of edge maps narrows down the domain gap between the edge images converted from color photographs (during training) and hand-drawn sketches (during inference).
Figure~\ref{fig:edge} shows examples of standardized edge maps.

\subsection{Sampling Process}
The Stable Diffusion provides text-guided image-to-image translation based on SDEdit~\cite{meng2021sdedit}.
This feature can control how much of an initial image is preserved by having a strength parameter ($[0.0, 1.0]$), \ie, it controls the amount of noise added to the initial image.
However, in our case, we want to preserve only the semantic content and shapes from sketches, not any visual features.
Therefore, we focus on full sampling from pure noise with sketch conditioning.
As in the original Stable Diffusion, we follow classifier-free guidance (see Eq.~\eqref{eq:cfg}) in this work.

\section{Experiments}
\subsection{Setup}

\noindent{\bf Dataset:}
We used improved aesthetics $6.5$+, which is a subset of the LAION dataset~\cite{schuhmann2022laion}, to fine-tune the Stable Diffusion models for sketch conditioning.
For evaluation, we utilized random 300 samples from GeoPose3K Mountain Landscape~\cite{brejcha2017geopose3k} and the validation set (300 samples) of LSUN-Church~\cite{yu2015lsun}.
Since there are 600 test samples which capture different scenes from two datasets, we use BLIP~\cite{li2022blip} for captioning the samples.
In addition, we collected 50 line (sketch) images from the LAION dataset for subjective evaluation.

\vspace{2mm}\noindent{\bf Implementation details:}
We followed the default settings of the original Stable Diffusion~\cite{rombach2022high}.
For training, all images were resized into a dimension of $512\times512$.
We fine-tuned depth-to-image (one-channel concatenation) for 110k steps and upscale models (three-channel concatenation) for 85k steps with a batch size of $32$.

\vspace{2mm}\noindent{\bf Metrics:}
To evaluate the quality of synthesized images, we calculated Learned Perceptual Image Patch Similarity (LPIPS)~\cite{zhang2018unreasonable} scores and compared the performance among the models.

\begin{figure*}[htb]
\centering
    \renewcommand{\arraystretch}{0.5}
    \setlength\tabcolsep{1pt}
    \resizebox{\linewidth}{!}{%
    \begin{tabular}{ccccc|ccccc|ccccc}
    & \multicolumn{3}{c}{\scriptsize GeoPose3K} & & & \multicolumn{3}{c}{\scriptsize LSUN-Church} & & & \multicolumn{4}{c}{\scriptsize Line Images}\\
    \parbox[t]{2mm}{\rotatebox[origin=c]{90}{\scriptsize Input}} & 
    \begin{minipage}{0.09\linewidth}\includegraphics[width=\linewidth]{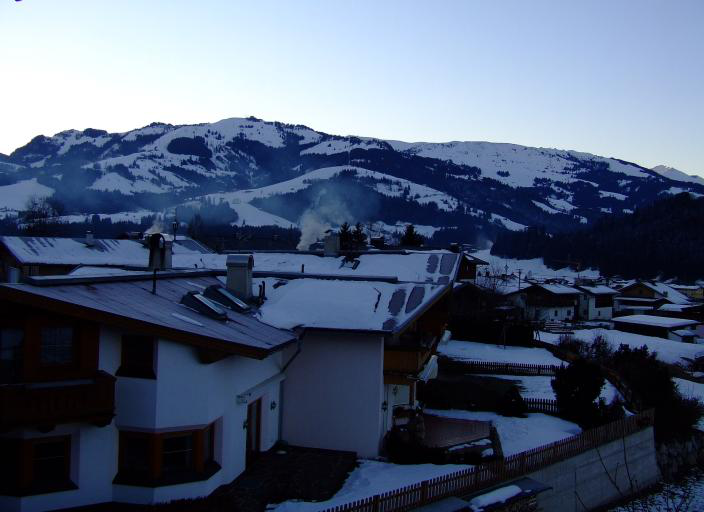}\end{minipage} & 
    \begin{minipage}{0.09\linewidth}\includegraphics[width=\linewidth]{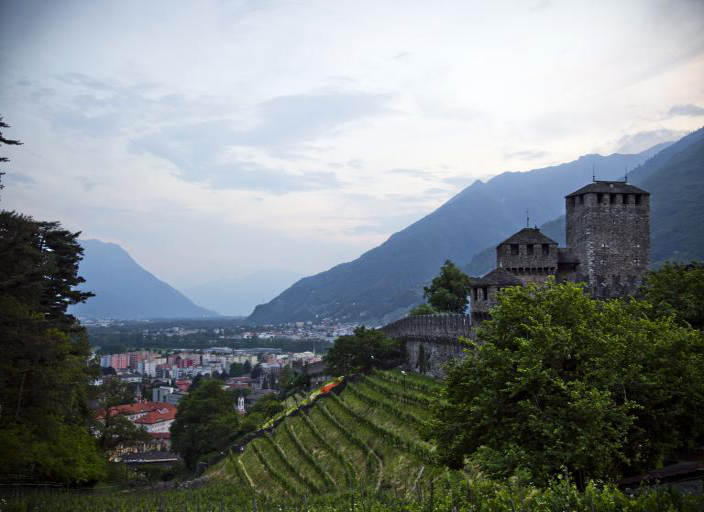}\end{minipage} &
    \begin{minipage}{0.09\linewidth}\includegraphics[width=\linewidth]{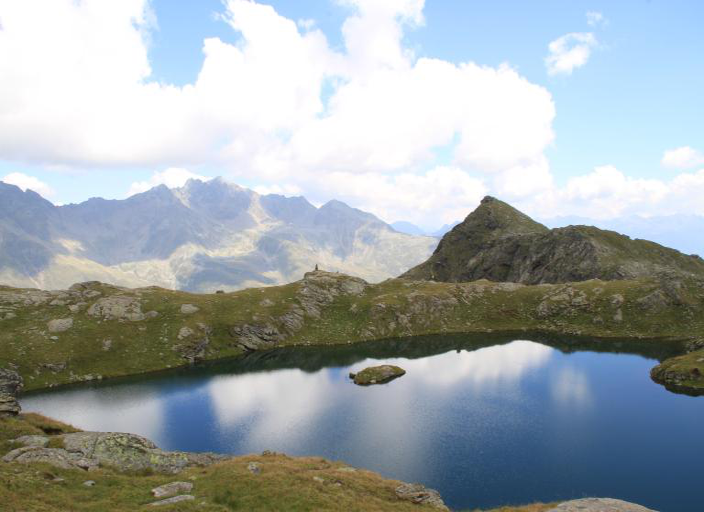}\end{minipage} & & &
    \begin{minipage}{0.09\linewidth}\includegraphics[width=\linewidth]{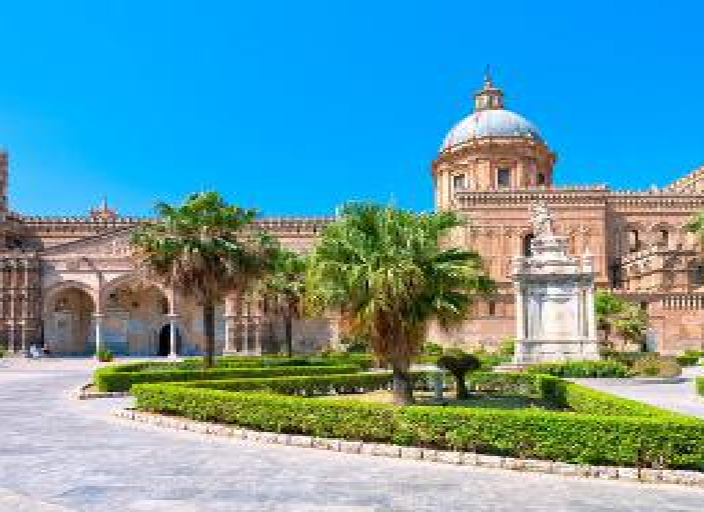}\end{minipage} &
    \begin{minipage}{0.09\linewidth}\includegraphics[width=\linewidth]{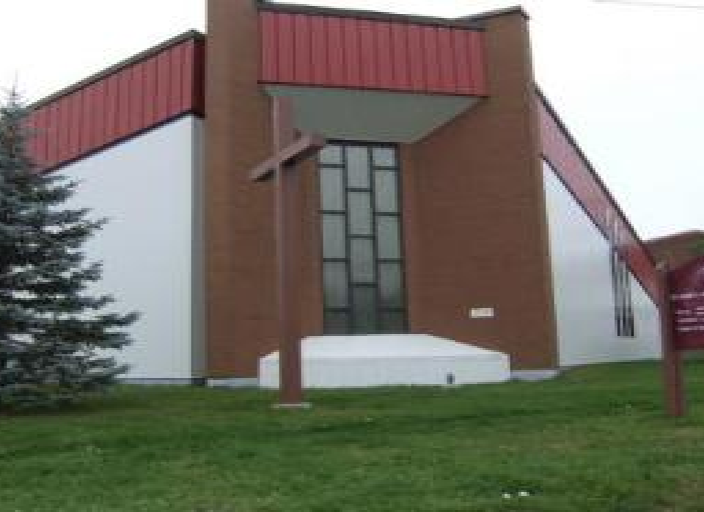}\end{minipage} &
    \begin{minipage}{0.09\linewidth}\includegraphics[width=\linewidth]{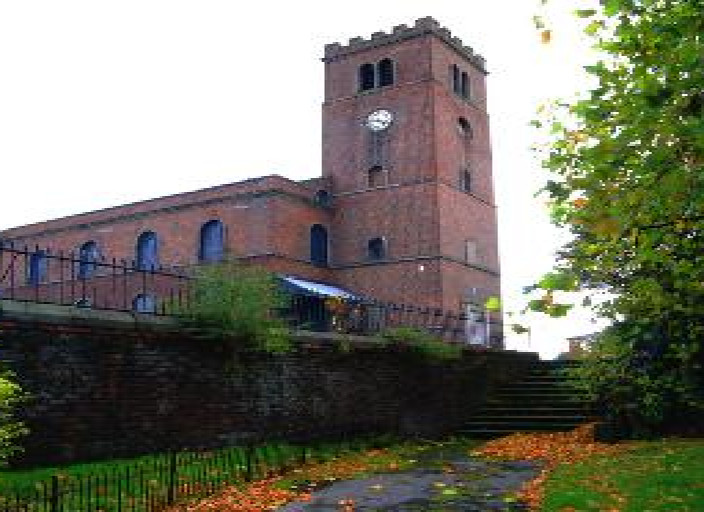}\end{minipage} & & &
    \begin{minipage}{0.09\linewidth}\includegraphics[width=\linewidth]{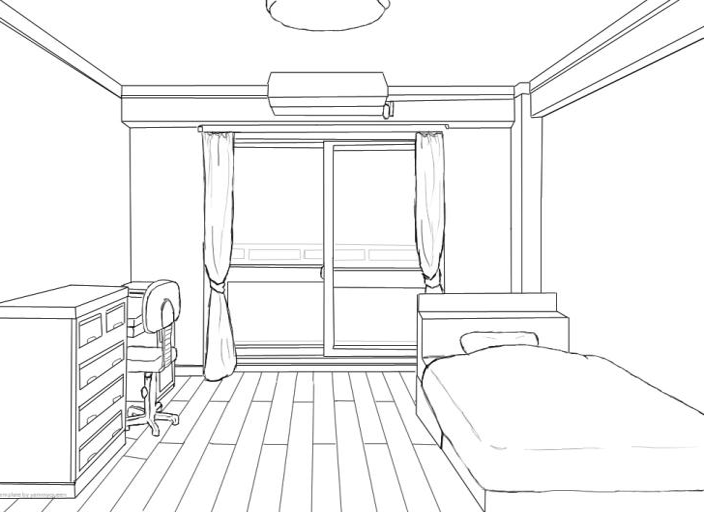}\end{minipage} &
    \begin{minipage}{0.09\linewidth}\includegraphics[width=\linewidth]{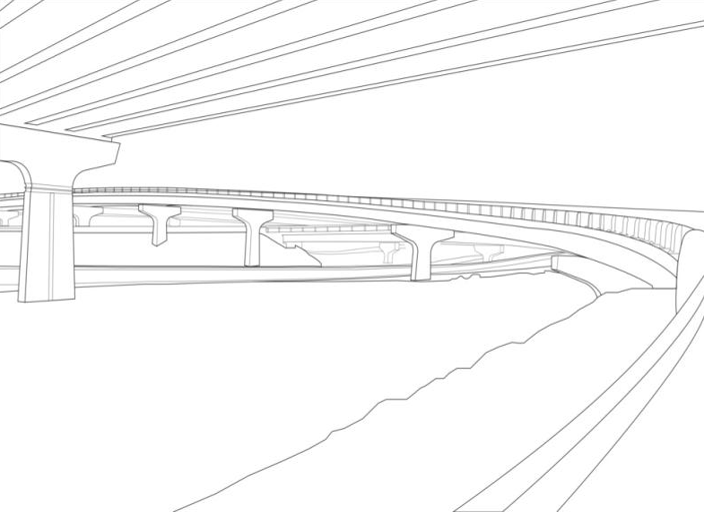}\end{minipage} &
    \begin{minipage}{0.09\linewidth}\includegraphics[width=\linewidth]{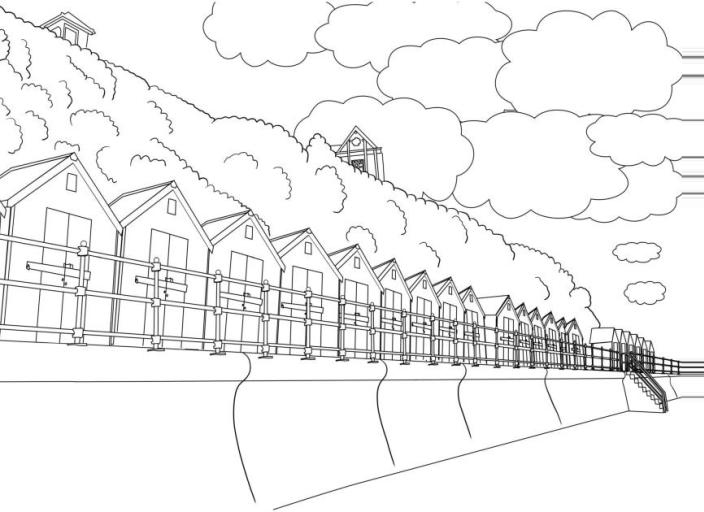}\end{minipage} &
    \begin{minipage}{0.09\linewidth}\includegraphics[width=\linewidth]{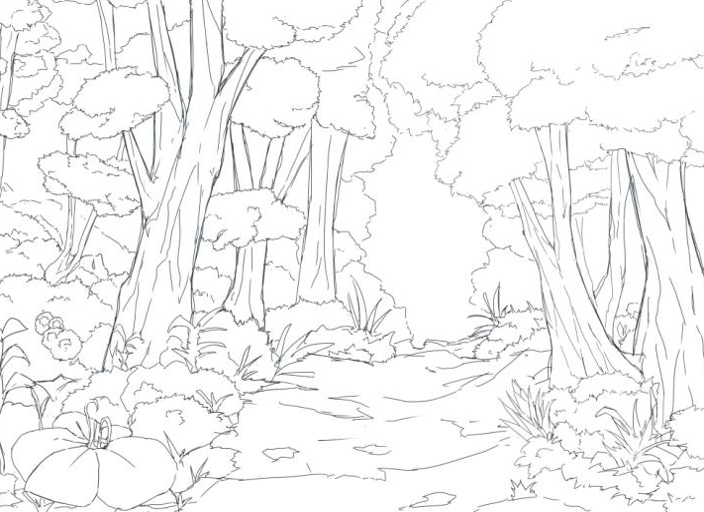}\end{minipage} \\[0.5cm]
    \parbox[t]{2mm}{\rotatebox[origin=c]{90}{\scriptsize Edge}} & 
    \begin{minipage}{0.09\linewidth}\includegraphics[width=\linewidth]{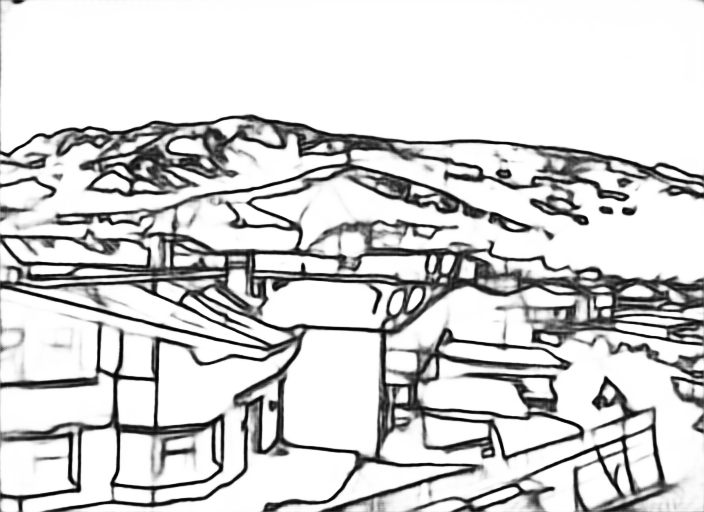}\end{minipage} & 
    \begin{minipage}{0.09\linewidth}\includegraphics[width=\linewidth]{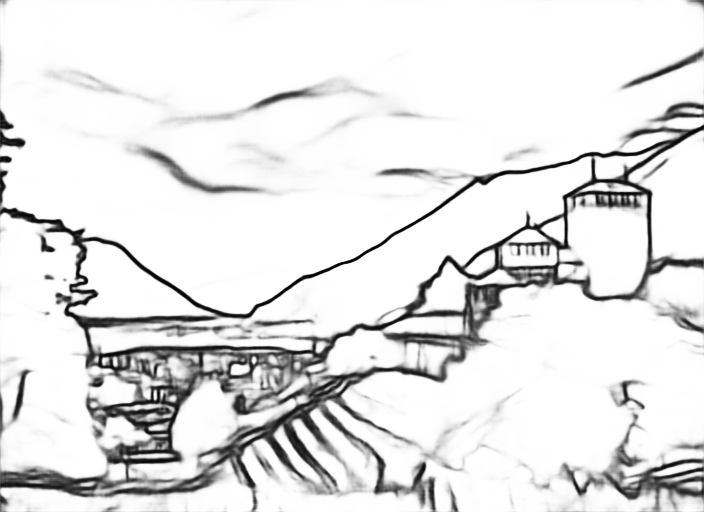}\end{minipage} &
    \begin{minipage}{0.09\linewidth}\includegraphics[width=\linewidth]{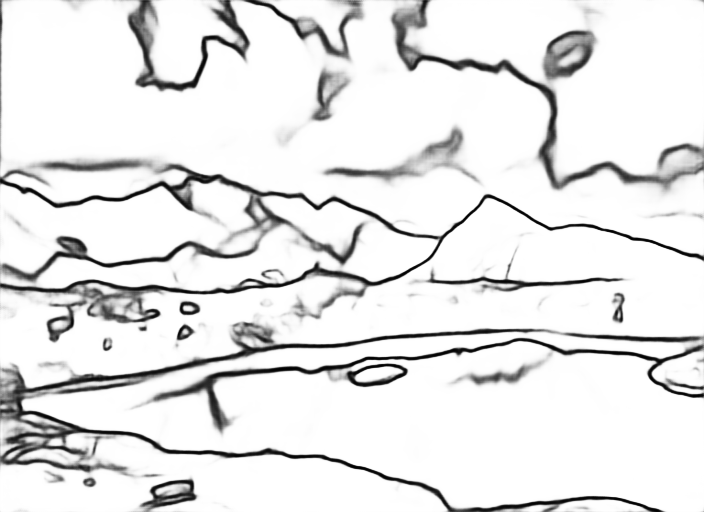}\end{minipage} & & &
    \begin{minipage}{0.09\linewidth}\includegraphics[width=\linewidth]{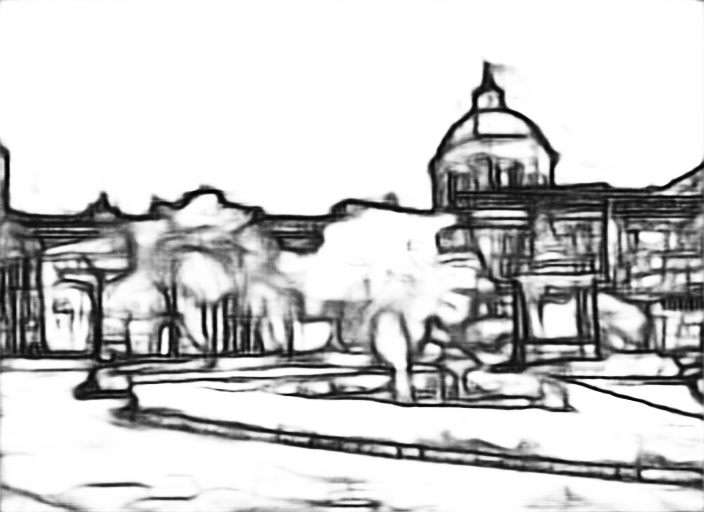}\end{minipage} &
    \begin{minipage}{0.09\linewidth}\includegraphics[width=\linewidth]{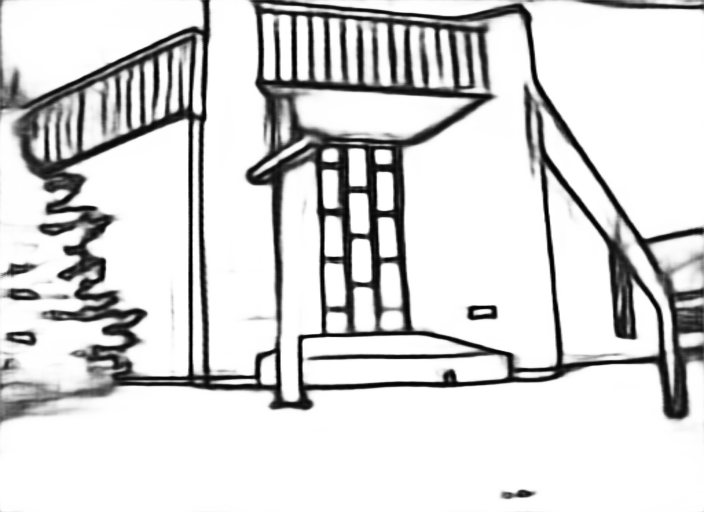}\end{minipage} &
    \begin{minipage}{0.09\linewidth}\includegraphics[width=\linewidth]{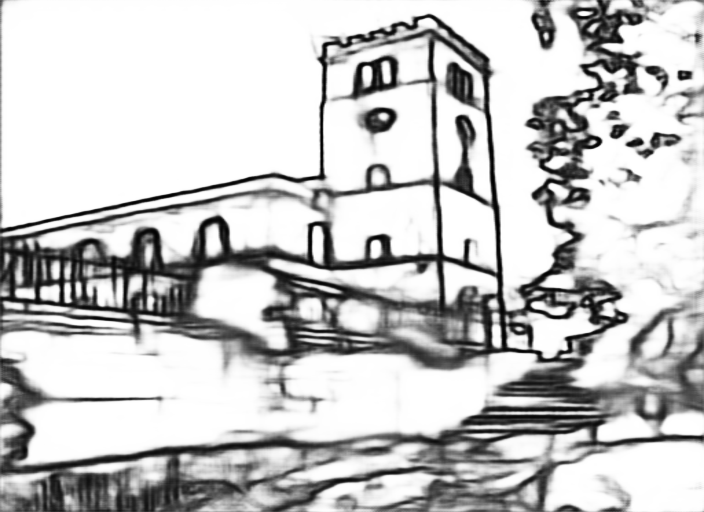}\end{minipage} & & &
    \begin{minipage}{0.09\linewidth}\includegraphics[width=\linewidth]{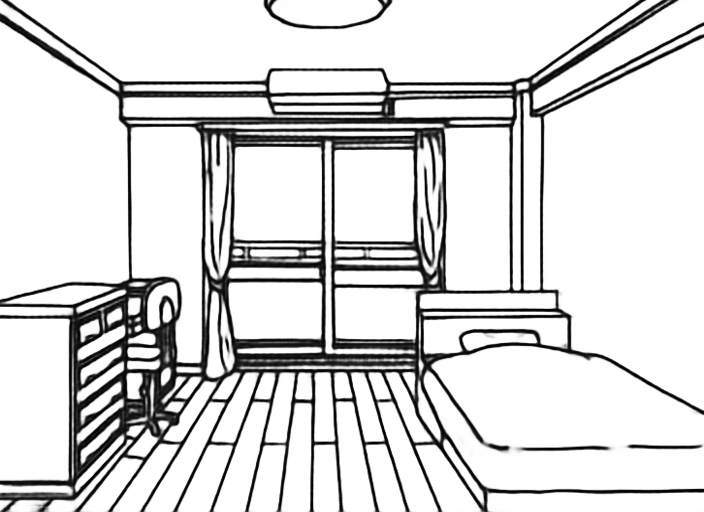}\end{minipage} &
    \begin{minipage}{0.09\linewidth}\includegraphics[width=\linewidth]{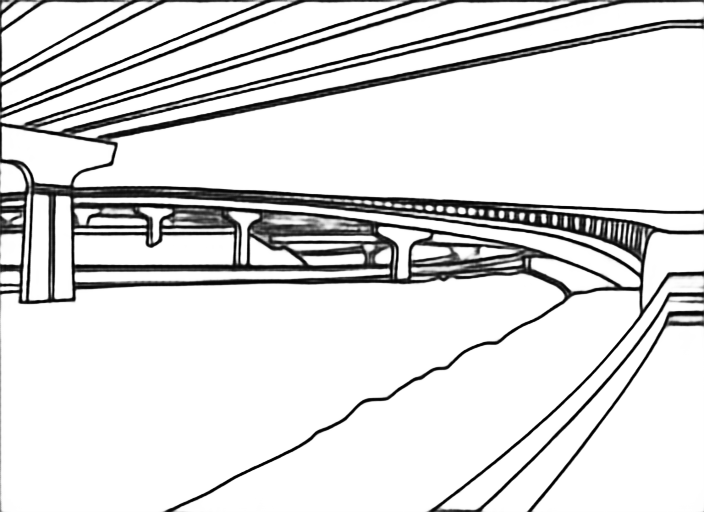}\end{minipage} &
    \begin{minipage}{0.09\linewidth}\includegraphics[width=\linewidth]{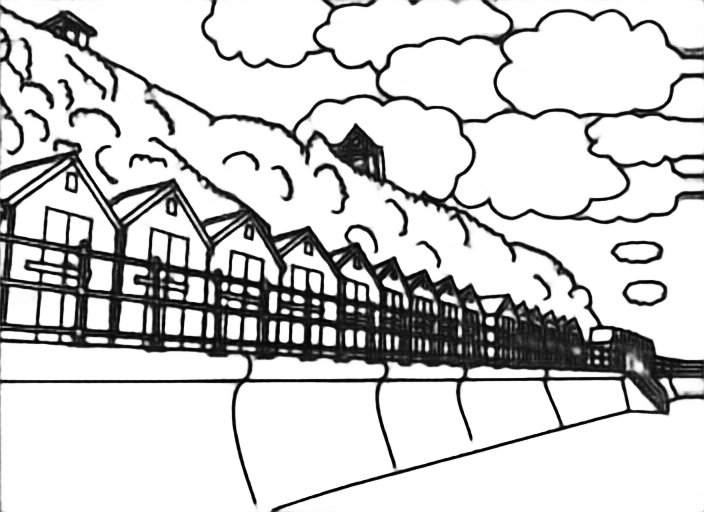}\end{minipage} &
    \begin{minipage}{0.09\linewidth}\includegraphics[width=\linewidth]{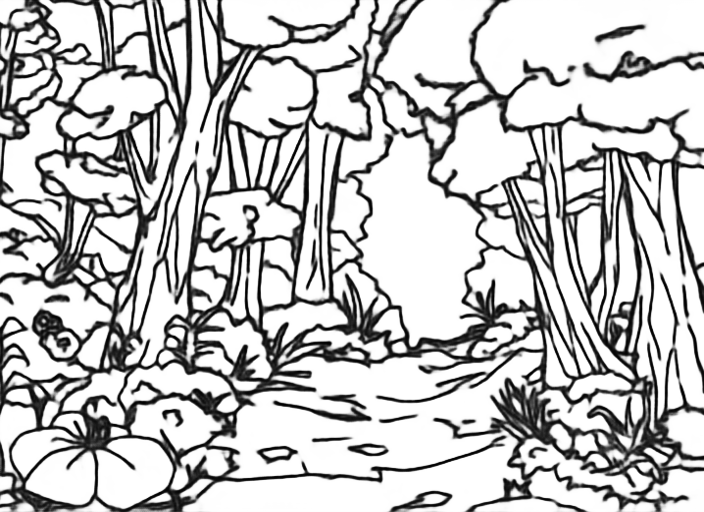}\end{minipage} \\[0.5cm]
    \parbox[t]{2mm}{\rotatebox[origin=c]{90}{\scriptsize Concat1}} & 
    \begin{minipage}{0.09\linewidth}\includegraphics[width=\linewidth]{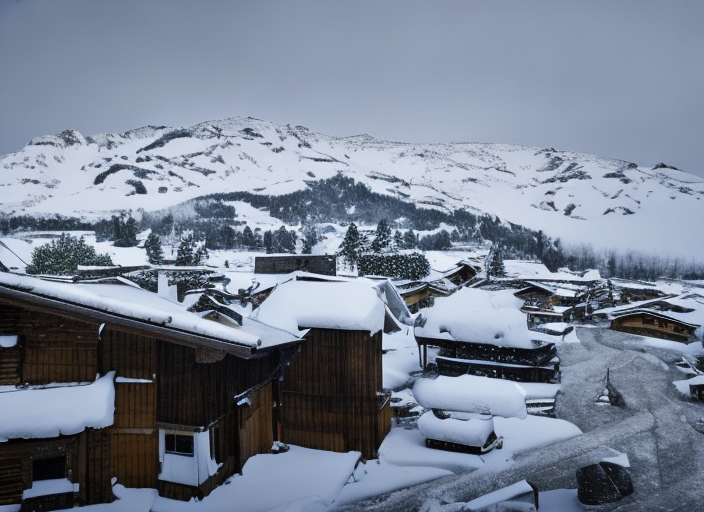}\end{minipage} & 
    \begin{minipage}{0.09\linewidth}\includegraphics[width=\linewidth]{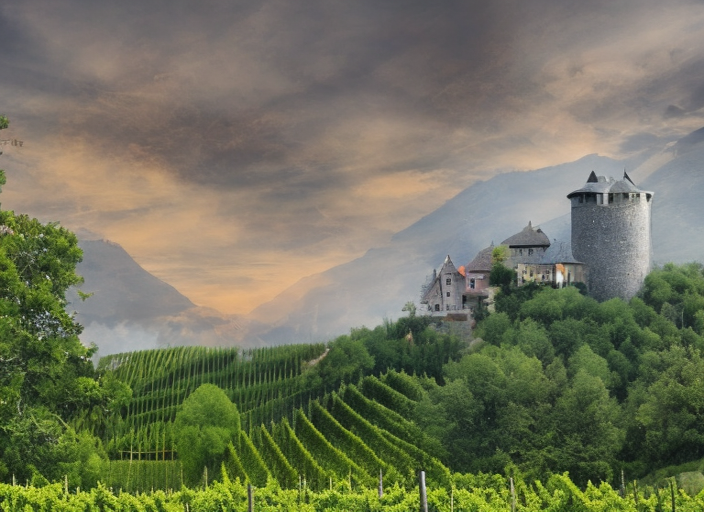}\end{minipage} &
    \begin{minipage}{0.09\linewidth}\includegraphics[width=\linewidth]{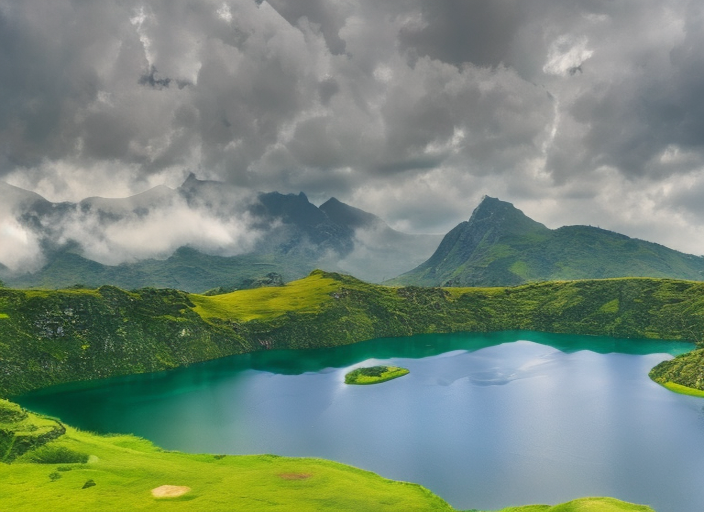}\end{minipage} & & &
    \begin{minipage}{0.09\linewidth}\includegraphics[width=\linewidth]{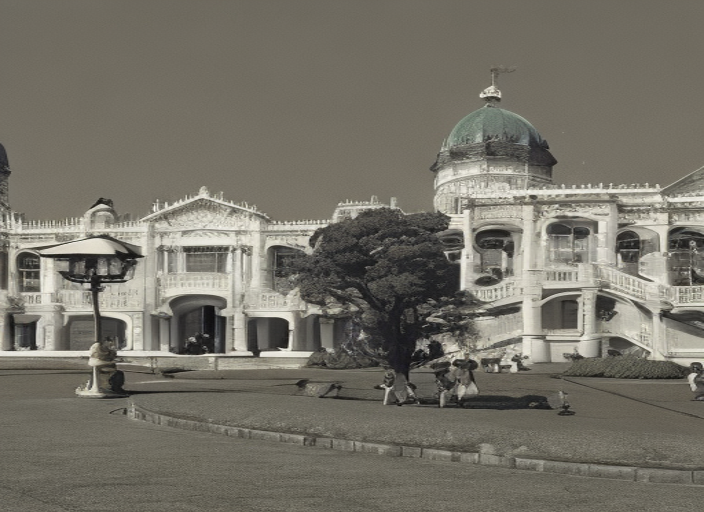}\end{minipage} &
    \begin{minipage}{0.09\linewidth}\includegraphics[width=\linewidth]{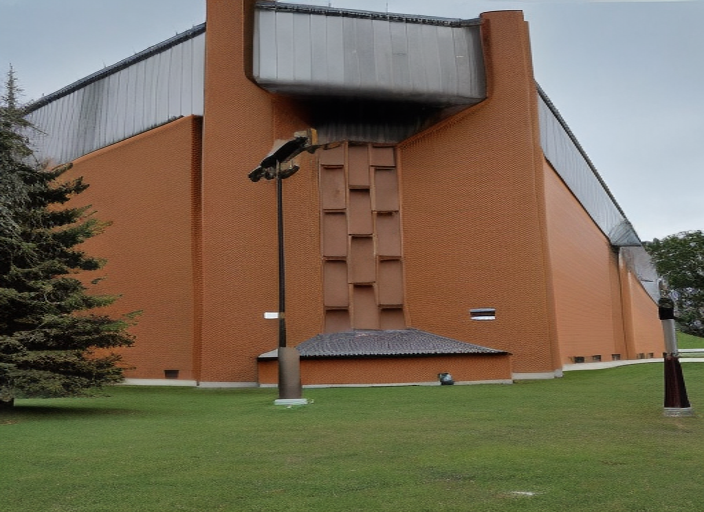}\end{minipage} &
    \begin{minipage}{0.09\linewidth}\includegraphics[width=\linewidth]{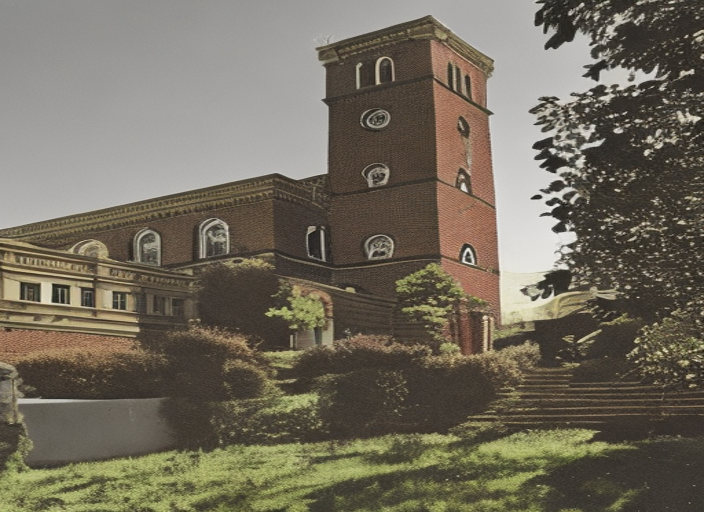}\end{minipage} & & &
    \begin{minipage}{0.09\linewidth}\includegraphics[width=\linewidth]{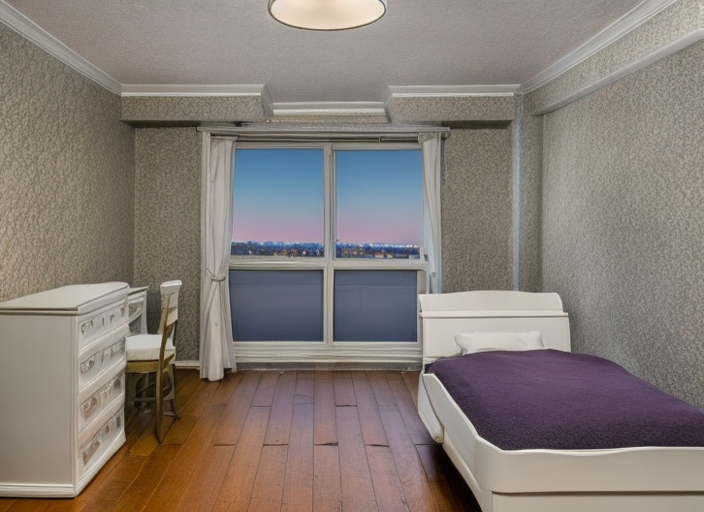}\end{minipage} &
    \begin{minipage}{0.09\linewidth}\includegraphics[width=\linewidth]{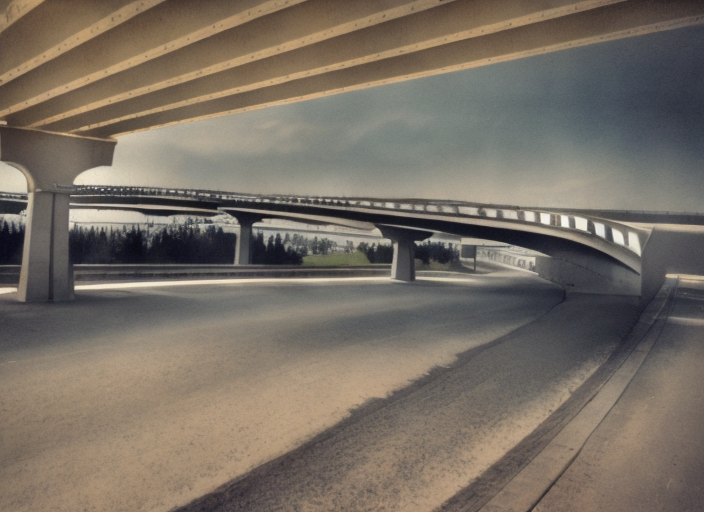}\end{minipage} &
    \begin{minipage}{0.09\linewidth}\includegraphics[width=\linewidth]{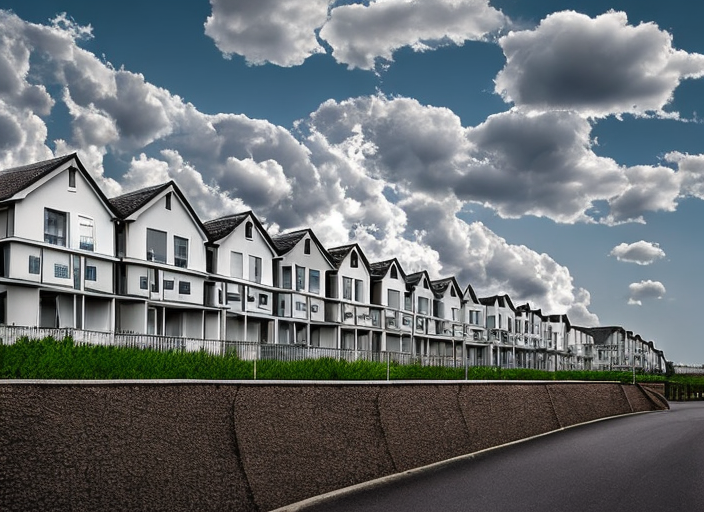}\end{minipage} &
    \begin{minipage}{0.09\linewidth}\includegraphics[width=\linewidth]{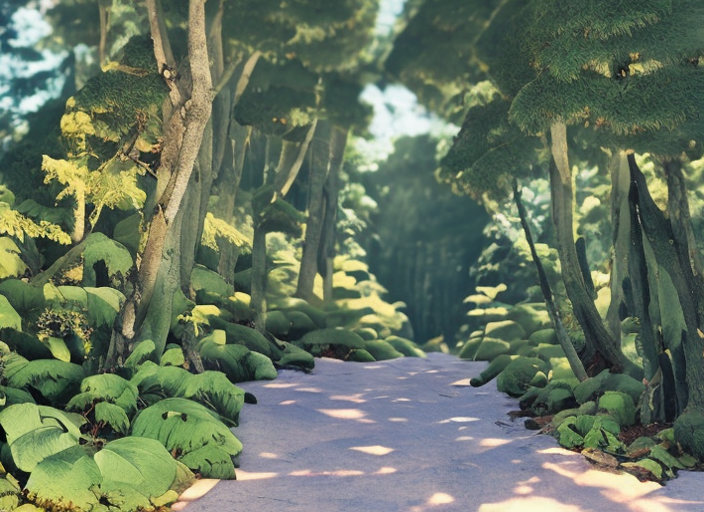}\end{minipage} \\[0.5cm]
    \parbox[t]{2mm}{\rotatebox[origin=c]{90}{\scriptsize Concat3}} & 
    \begin{minipage}{0.09\linewidth}\includegraphics[width=\linewidth]{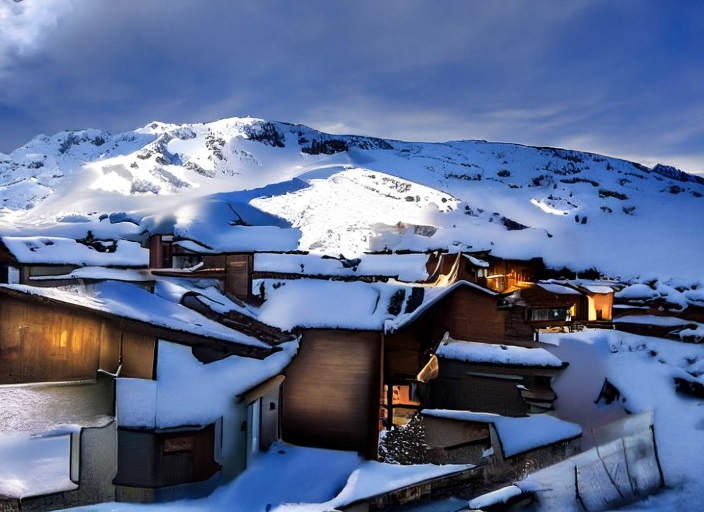}\end{minipage} & 
    \begin{minipage}{0.09\linewidth}\includegraphics[width=\linewidth]{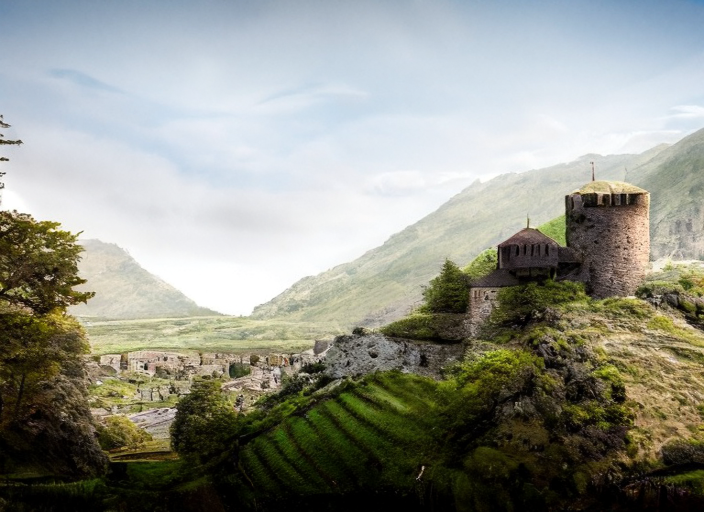}\end{minipage} &
    \begin{minipage}{0.09\linewidth}\includegraphics[width=\linewidth]{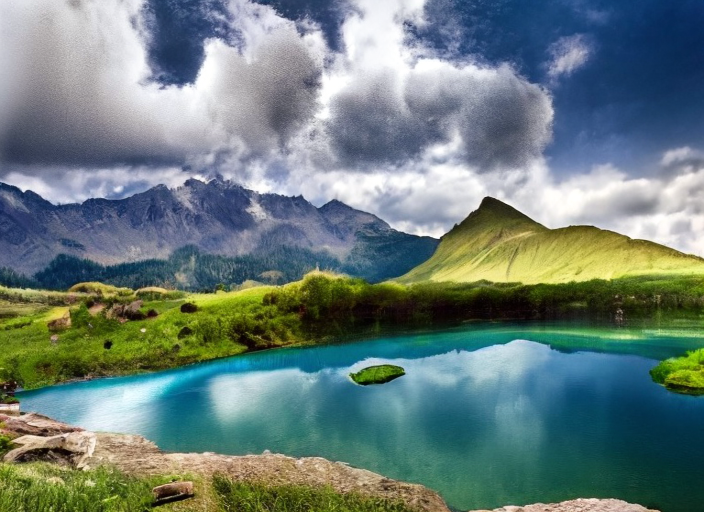}\end{minipage} & & &
    \begin{minipage}{0.09\linewidth}\includegraphics[width=\linewidth]{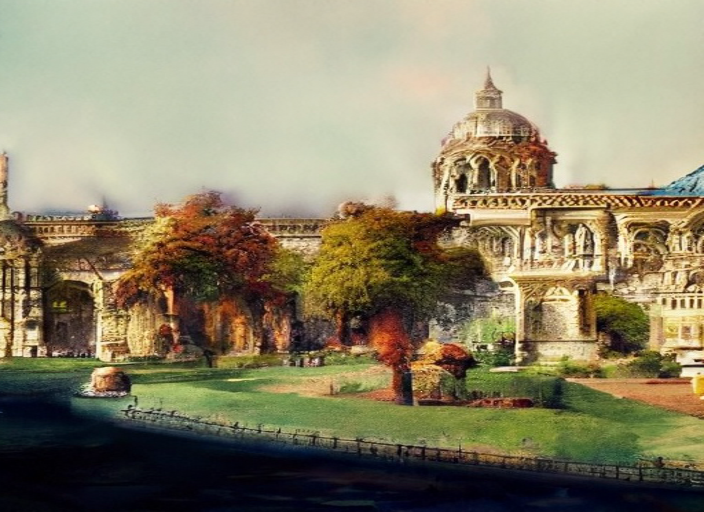}\end{minipage} &
    \begin{minipage}{0.09\linewidth}\includegraphics[width=\linewidth]{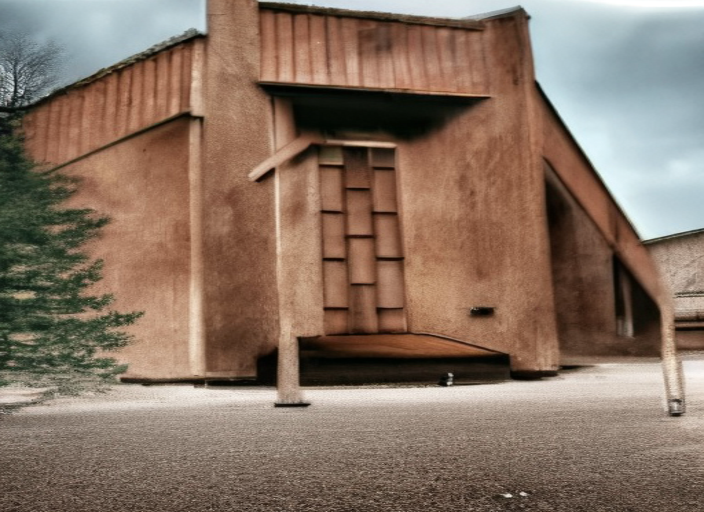}\end{minipage} &
    \begin{minipage}{0.09\linewidth}\includegraphics[width=\linewidth]{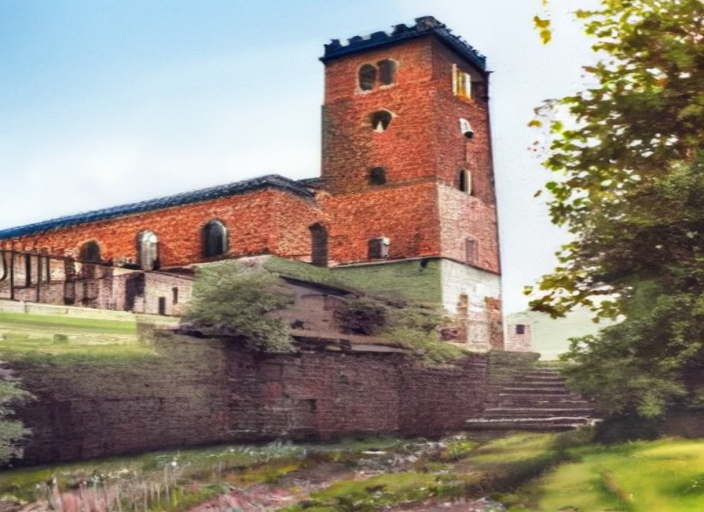}\end{minipage} & & &
    \begin{minipage}{0.09\linewidth}\includegraphics[width=\linewidth]{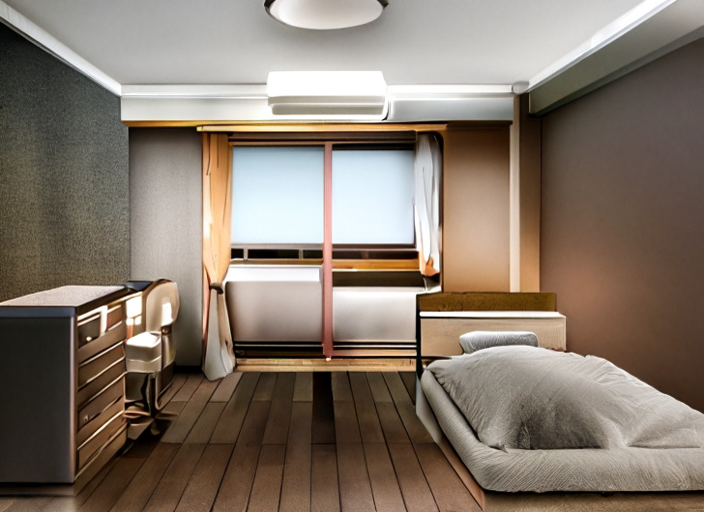}\end{minipage} &
    \begin{minipage}{0.09\linewidth}\includegraphics[width=\linewidth]{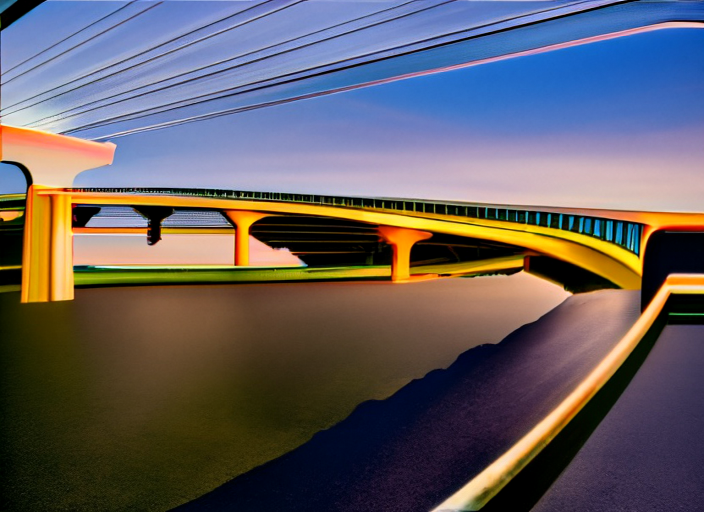}\end{minipage} &
    \begin{minipage}{0.09\linewidth}\includegraphics[width=\linewidth]{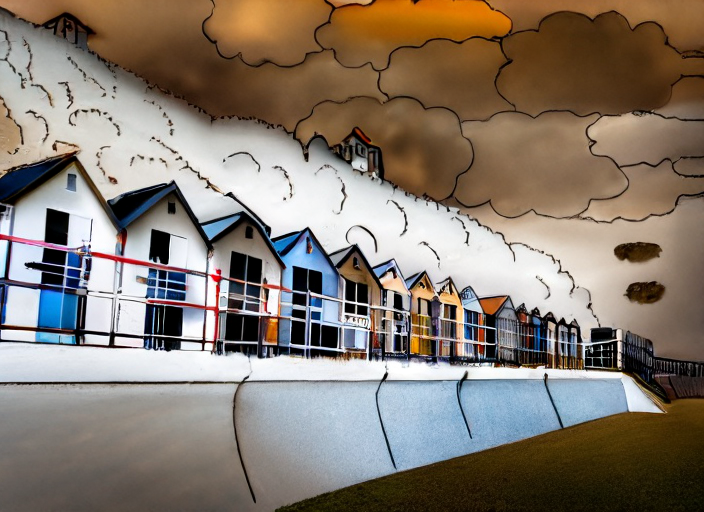}\end{minipage} &
    \begin{minipage}{0.09\linewidth}\includegraphics[width=\linewidth]{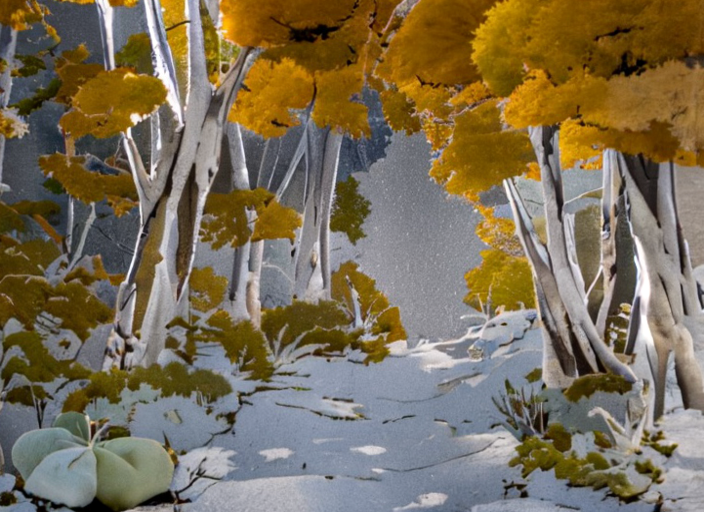}\end{minipage} \\[0.5cm]
    \parbox[t]{2mm}{\rotatebox[origin=c]{90}{\scriptsize PITI~\cite{wang2022pretraining}}} & 
    \begin{minipage}{0.09\linewidth}\includegraphics[width=\linewidth]{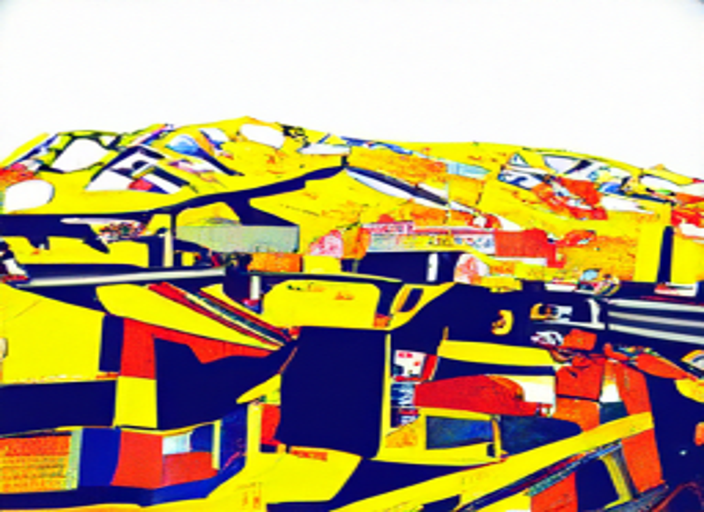}\end{minipage} & 
    \begin{minipage}{0.09\linewidth}\includegraphics[width=\linewidth]{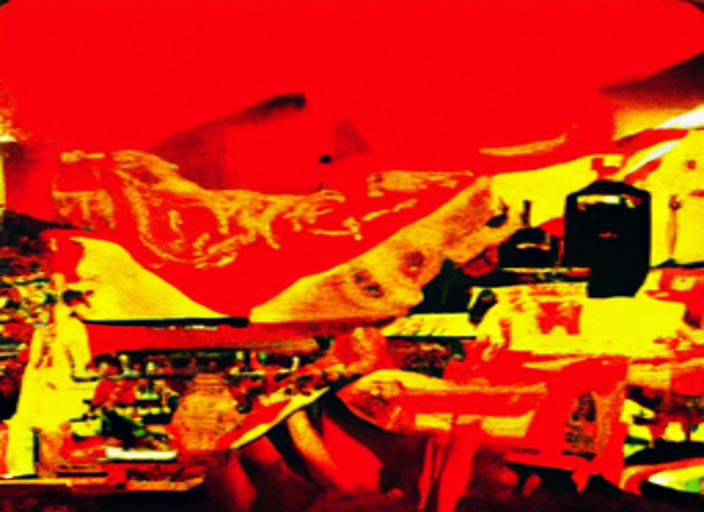}\end{minipage} &
    \begin{minipage}{0.09\linewidth}\includegraphics[width=\linewidth]{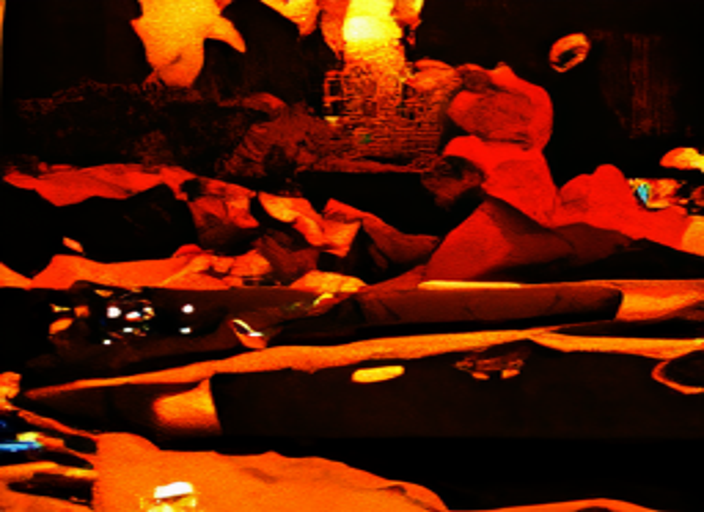}\end{minipage} & & &
    \begin{minipage}{0.09\linewidth}\includegraphics[width=\linewidth]{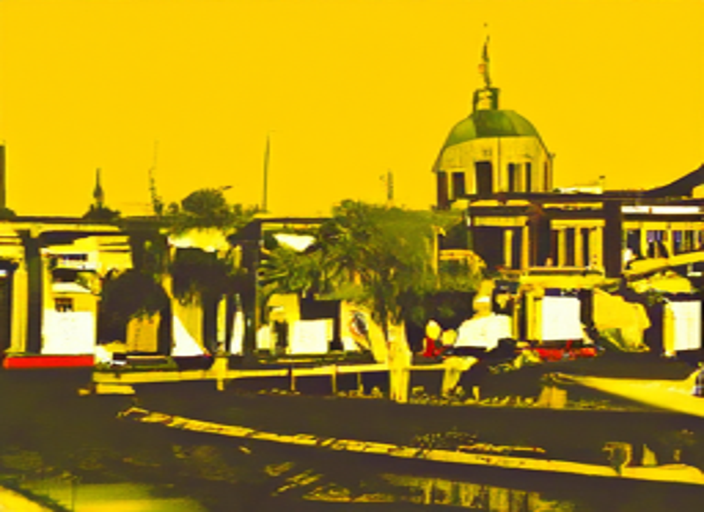}\end{minipage} &
    \begin{minipage}{0.09\linewidth}\includegraphics[width=\linewidth]{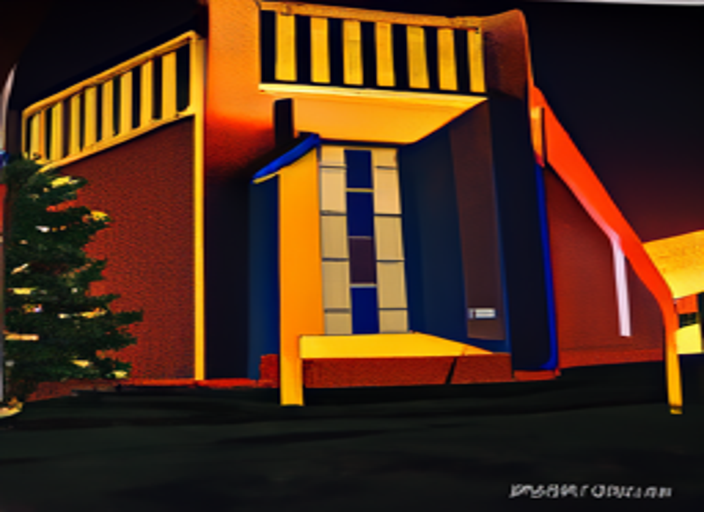}\end{minipage} &
    \begin{minipage}{0.09\linewidth}\includegraphics[width=\linewidth]{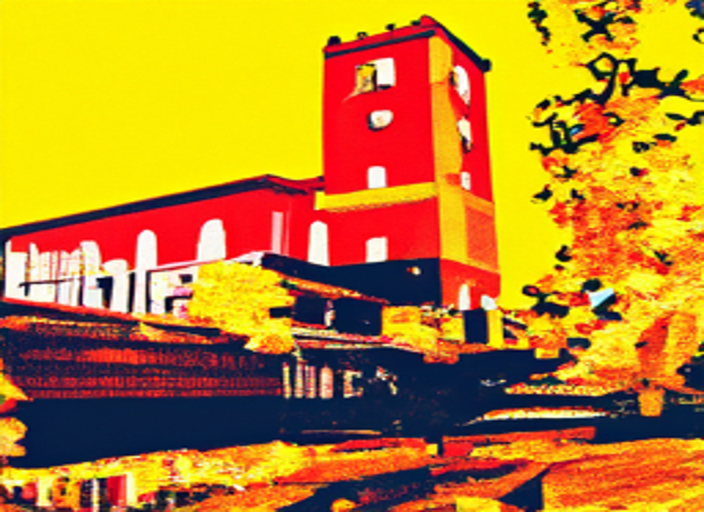}\end{minipage} & & &
    \begin{minipage}{0.09\linewidth}\includegraphics[width=\linewidth]{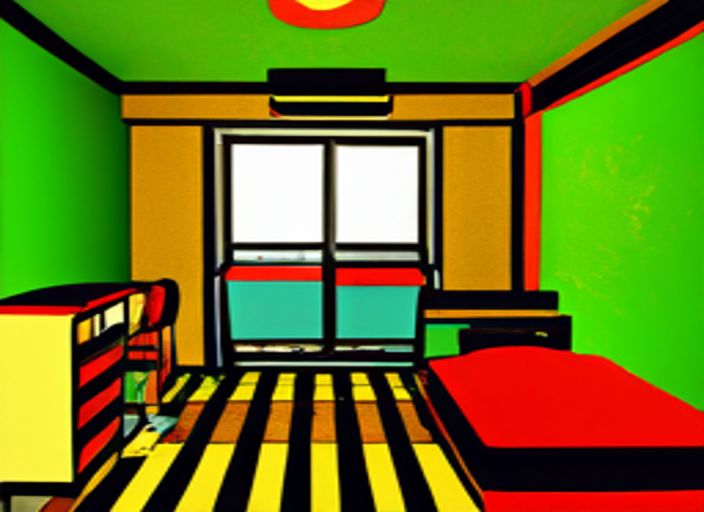}\end{minipage} &
    \begin{minipage}{0.09\linewidth}\includegraphics[width=\linewidth]{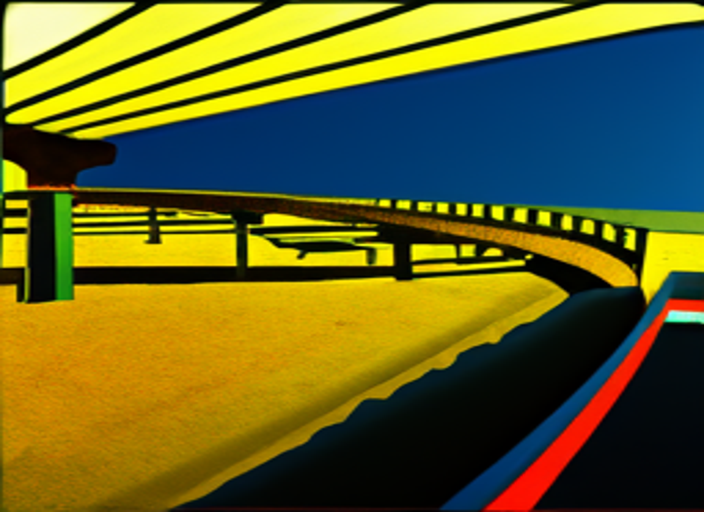}\end{minipage} &
    \begin{minipage}{0.09\linewidth}\includegraphics[width=\linewidth]{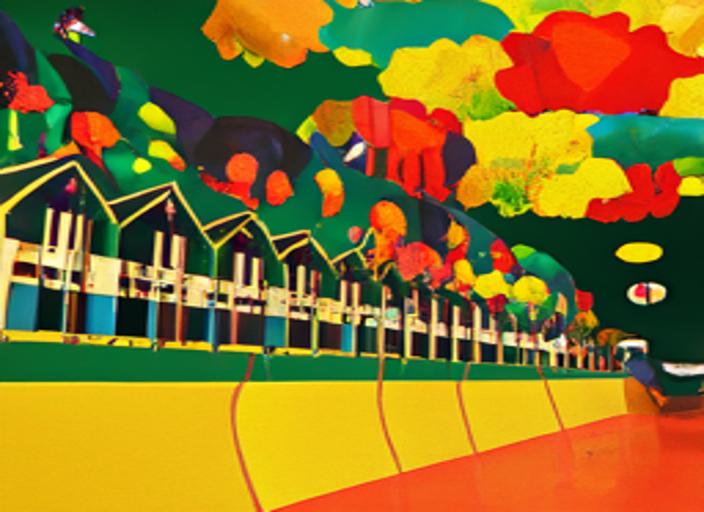}\end{minipage} &
    \begin{minipage}{0.09\linewidth}\includegraphics[width=\linewidth]{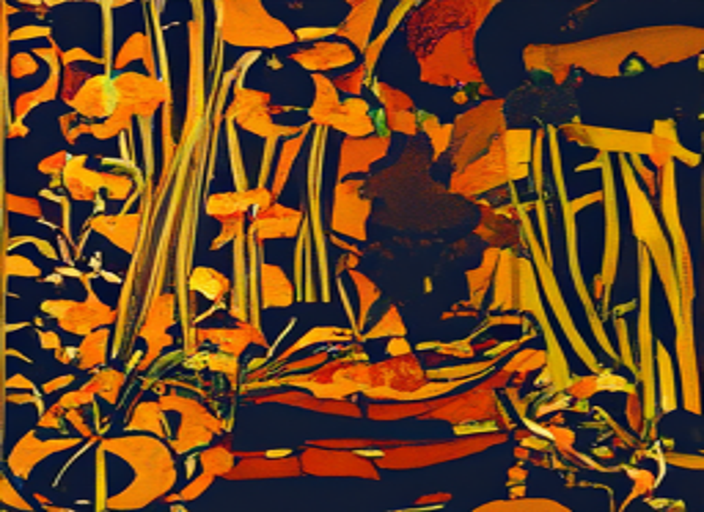}\end{minipage} \\
    \end{tabular}
    }
    \vspace{-.1in}
    \caption{Visual comparison of synthesized images. ``Concat1'' stands for one-channel concatenation model and ``Concat3'' is for three-channel concatenation model.\label{fig:comparison}}
\end{figure*}

\begin{figure*}[htb]
\centering
    \renewcommand{\arraystretch}{0.5}
    \setlength\tabcolsep{1pt}
    \resizebox{\linewidth}{!}{%
    \begin{tabular}{>{\centering}p{0.09\linewidth}>{\raggedright}p{0.09\linewidth}>{\raggedright}p{0.09\linewidth}>{\raggedright}p{0.09\linewidth}>{\raggedright}p{0.09\linewidth}>{\raggedright}p{0.09\linewidth}>{\raggedright}p{0.09\linewidth}>{\raggedright}p{0.09\linewidth}>{\raggedright}p{0.09\linewidth}>{\raggedright\arraybackslash}p{0.09\linewidth}}
    \scriptsize{Input} &
    \scriptsize{\emph{``an anime scene of [...]''}} & \scriptsize{\emph{``a watercolor painting of [...]''}} &
    \scriptsize{\emph{``an ukiyo-e art of [...]''}} & \scriptsize{\emph{``a black and white photograph of [...]''}} &
    \scriptsize{\emph{``a fresco painting of [...]''}} & \scriptsize{\emph{``a graffiti of [...]''}} &
    \scriptsize{\emph{``an oil painting of [...]''}} & \scriptsize{\emph{``a pop art of [...]''}} & \scriptsize{\emph{``an abstract art of [...]''}}\\
    \includegraphics[width=\linewidth]{line-input-1}&
    \includegraphics[width=\linewidth]{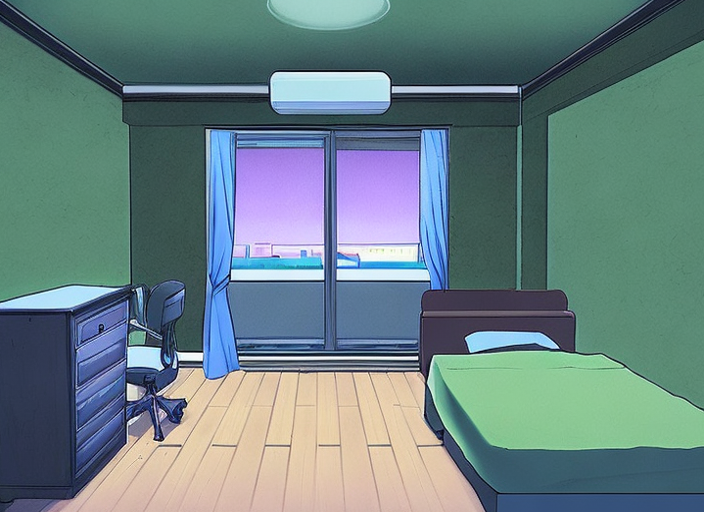}&
    \includegraphics[width=\linewidth]{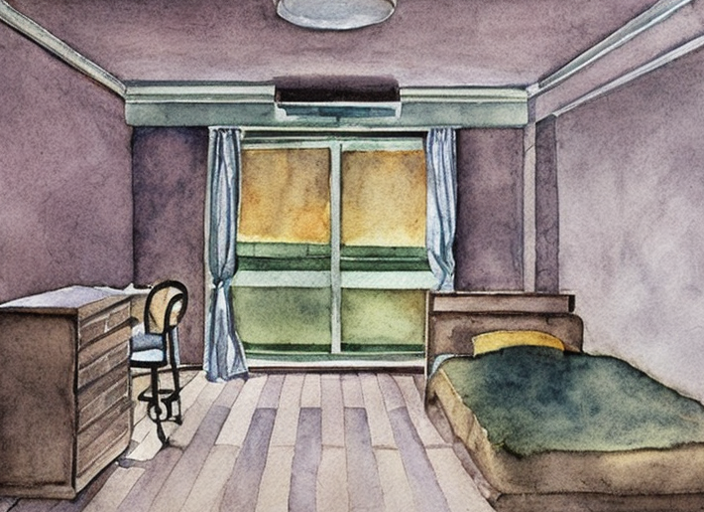}&
    \includegraphics[width=\linewidth]{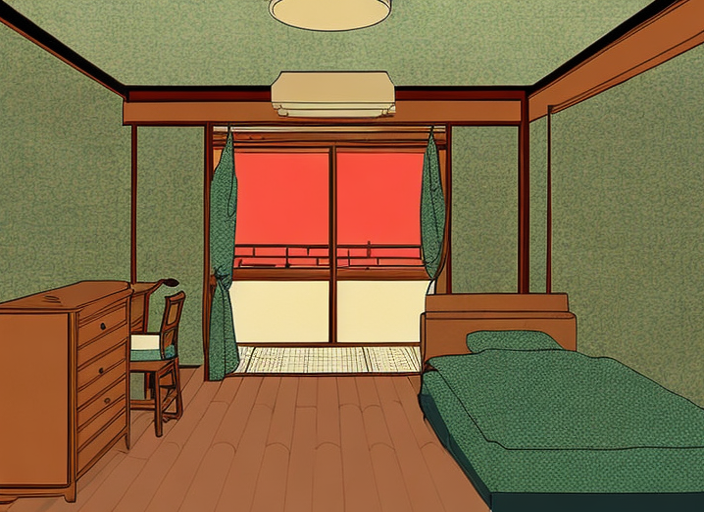}&
    \includegraphics[width=\linewidth]{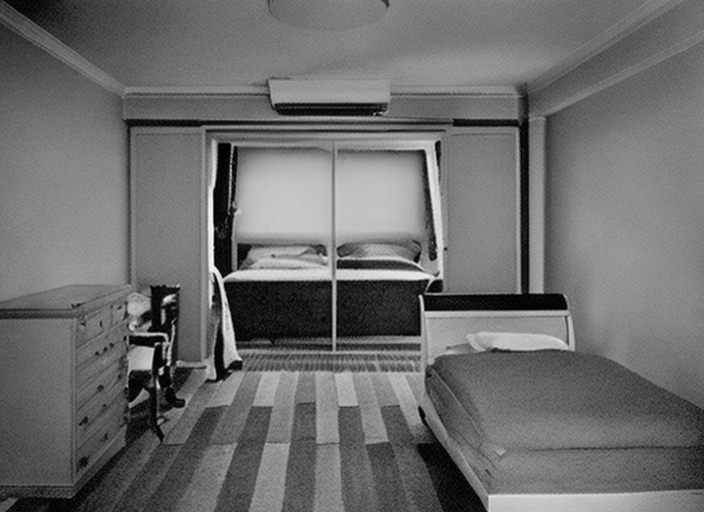}&
    \includegraphics[width=\linewidth]{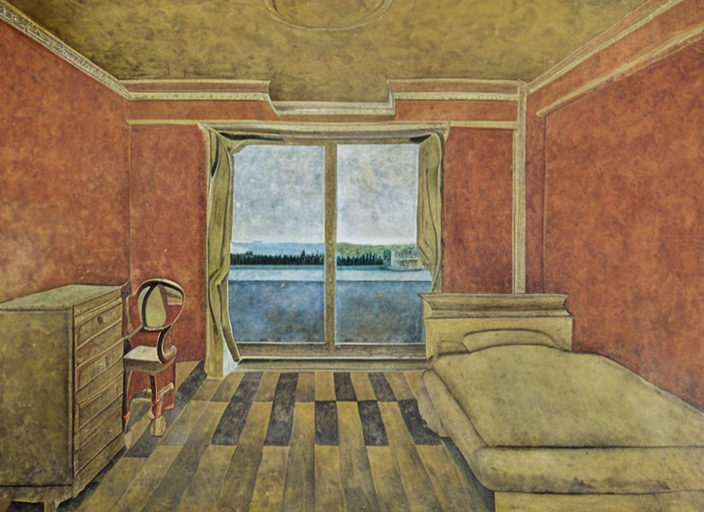}&
    \includegraphics[width=\linewidth]{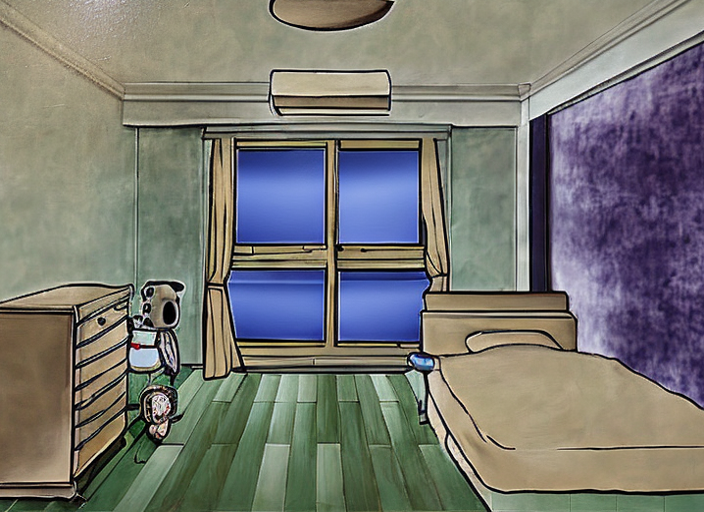}&
    \includegraphics[width=\linewidth]{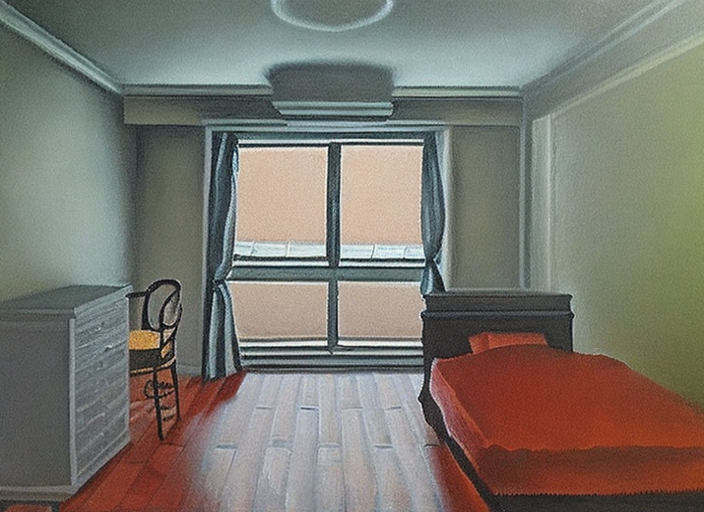}&
    \includegraphics[width=\linewidth]{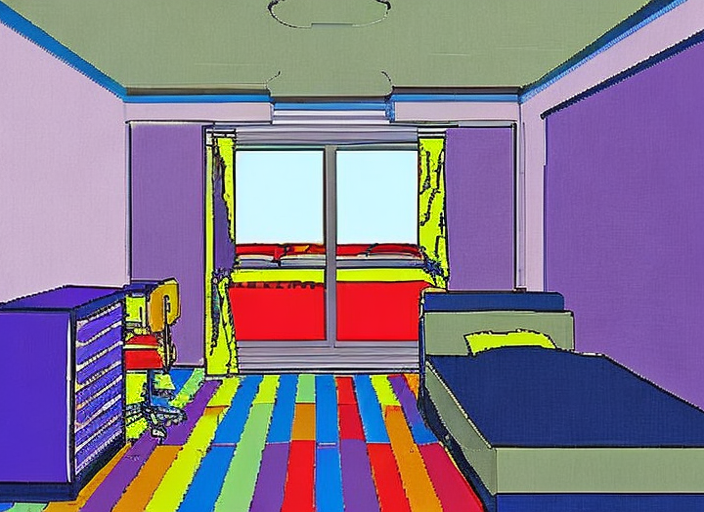}&
    \includegraphics[width=\linewidth]{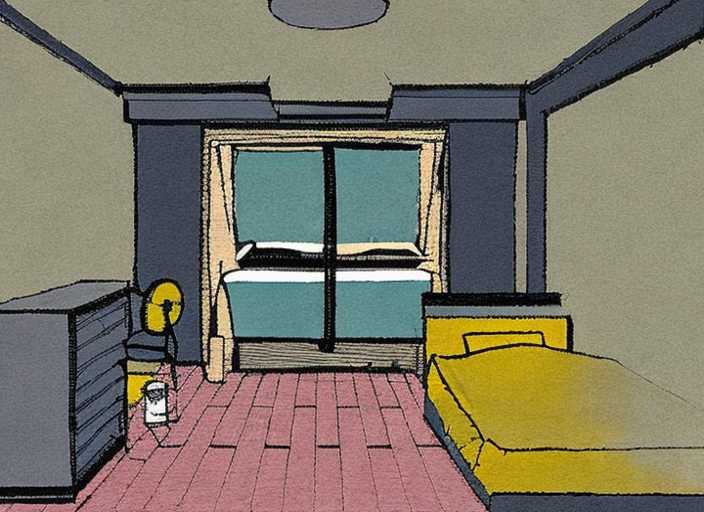}\\
    \includegraphics[width=\linewidth]{line-input-2}&
    \includegraphics[width=\linewidth]{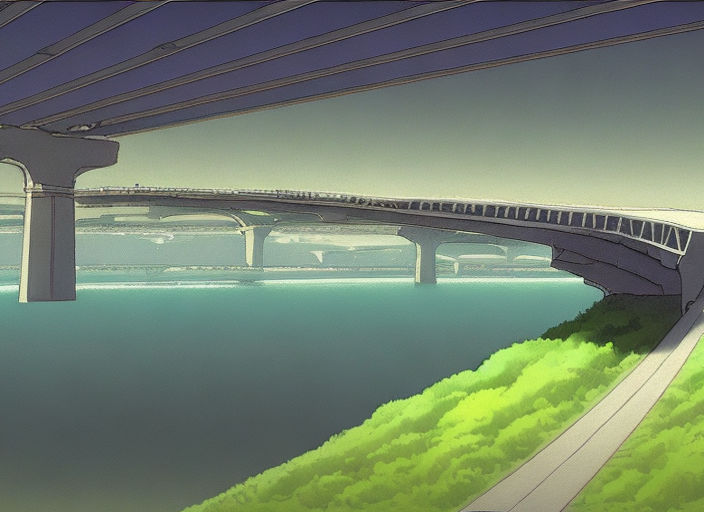}&
    \includegraphics[width=\linewidth]{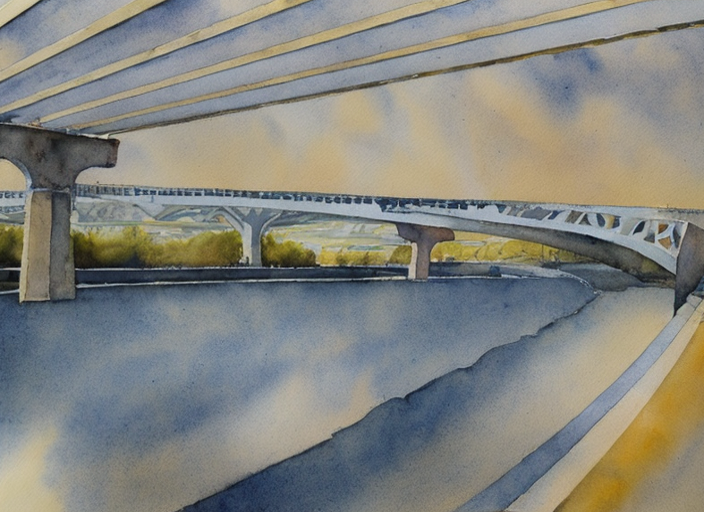}&
    \includegraphics[width=\linewidth]{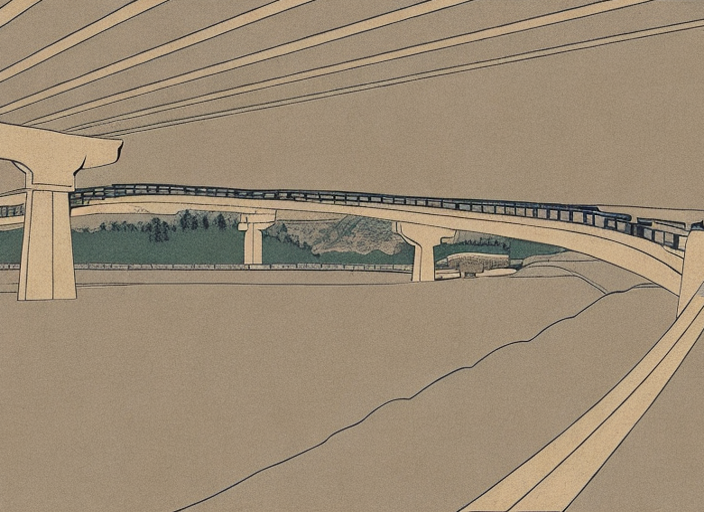}&
    \includegraphics[width=\linewidth]{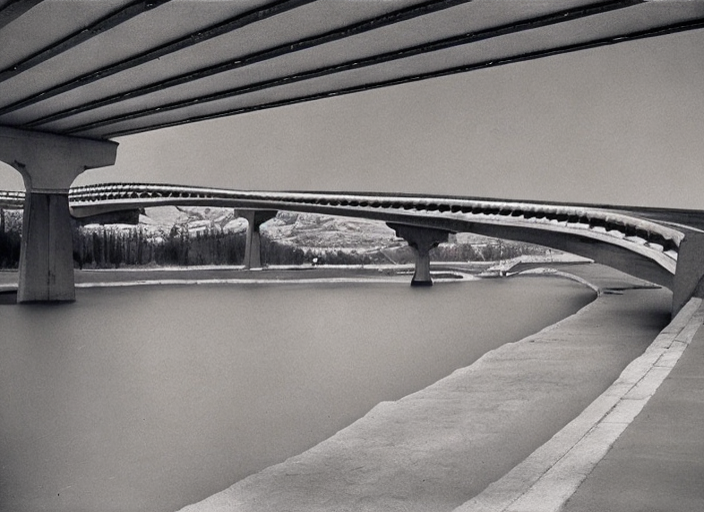}&
    \includegraphics[width=\linewidth]{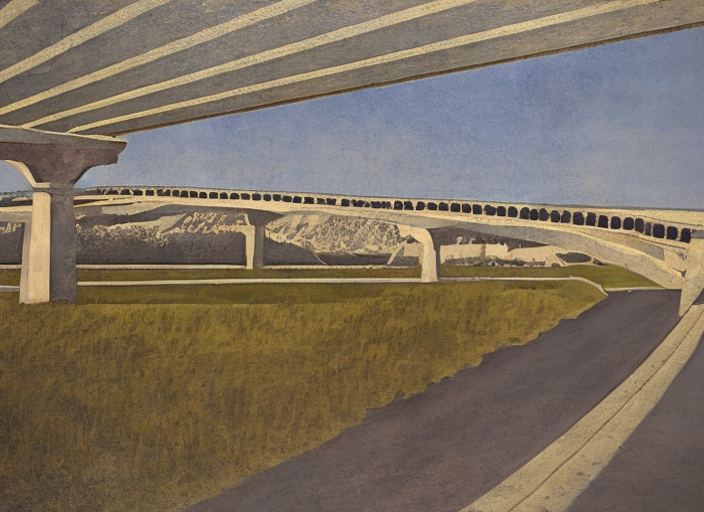}&
    \includegraphics[width=\linewidth]{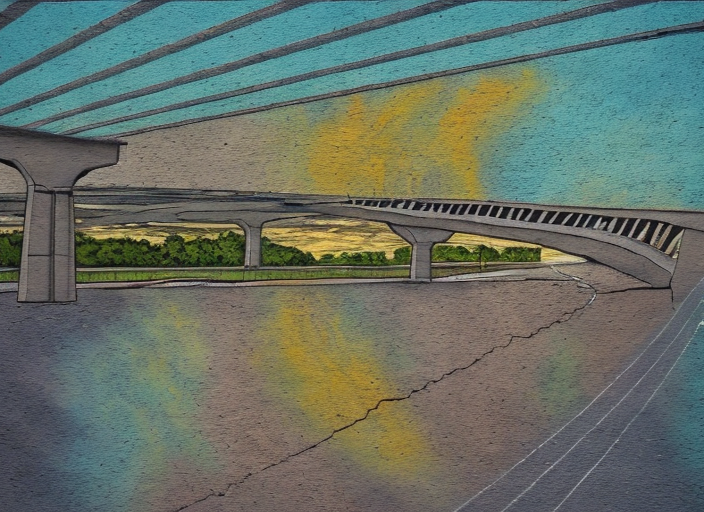}&
    \includegraphics[width=\linewidth]{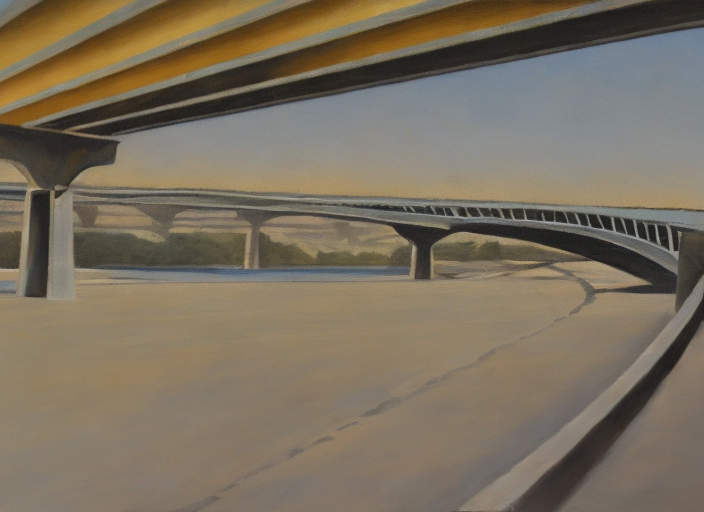}&
    \includegraphics[width=\linewidth]{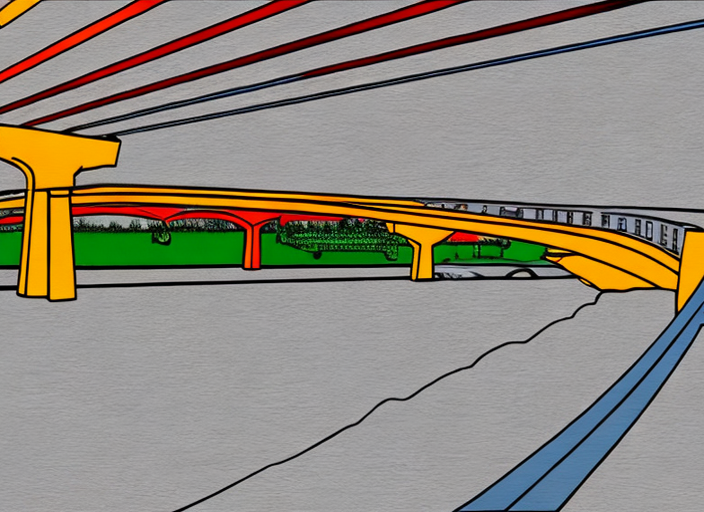}&
    \includegraphics[width=\linewidth]{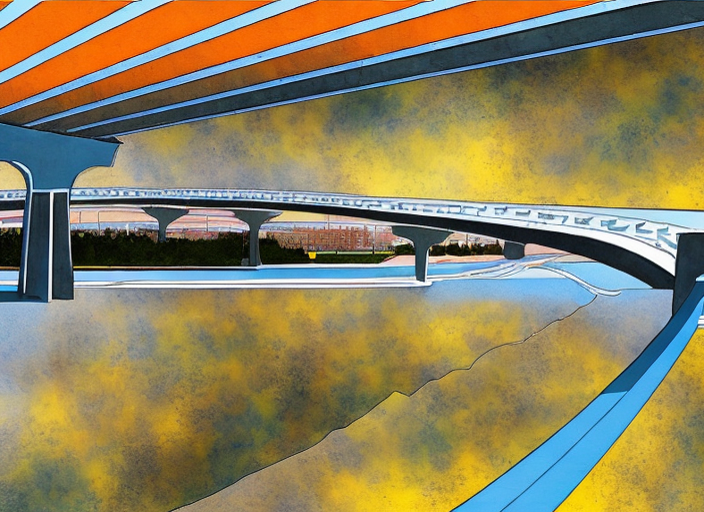}\\
    \includegraphics[width=\linewidth]{line-input-3}&
    \includegraphics[width=\linewidth]{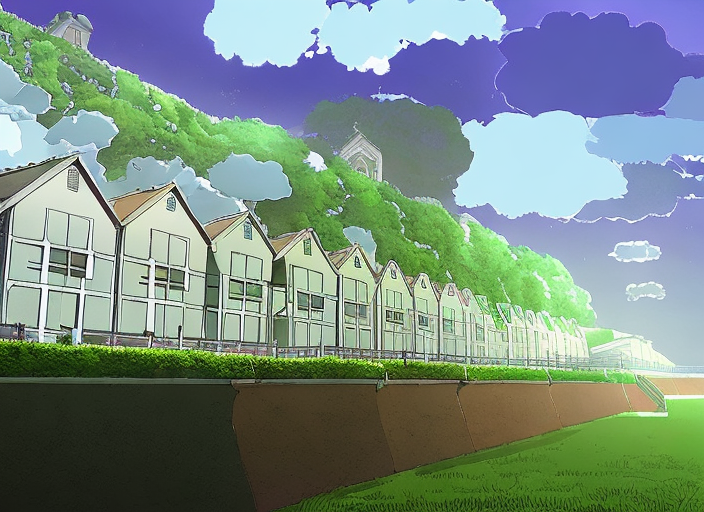}&
    \includegraphics[width=\linewidth]{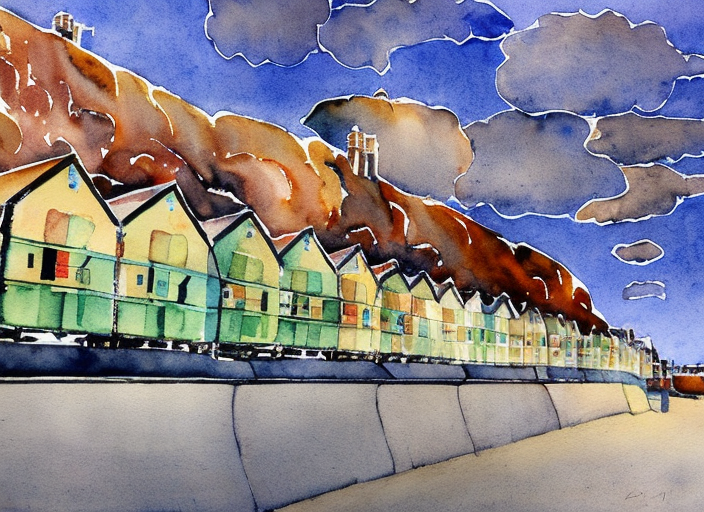}&
    \includegraphics[width=\linewidth]{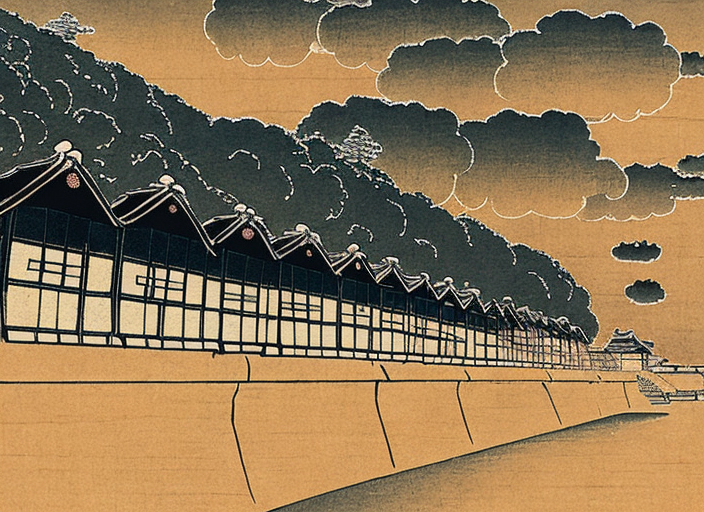}&
    \includegraphics[width=\linewidth]{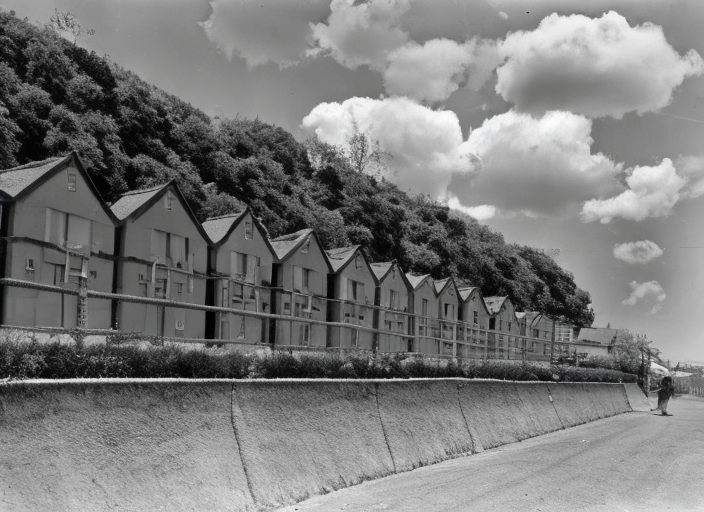}&
    \includegraphics[width=\linewidth]{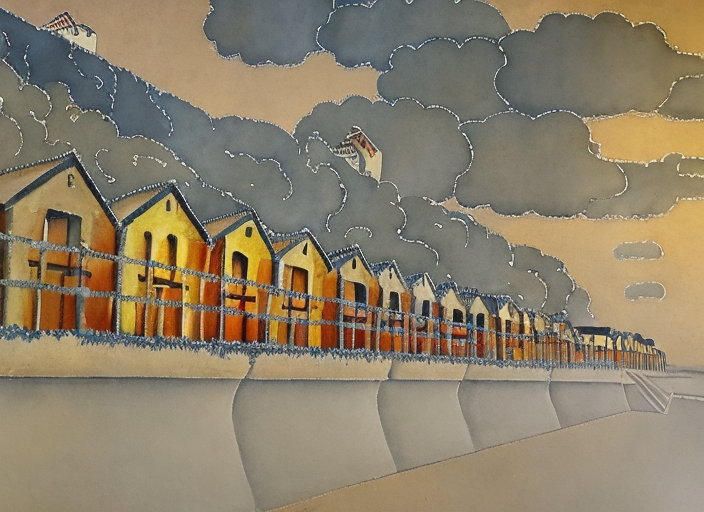}&
    \includegraphics[width=\linewidth]{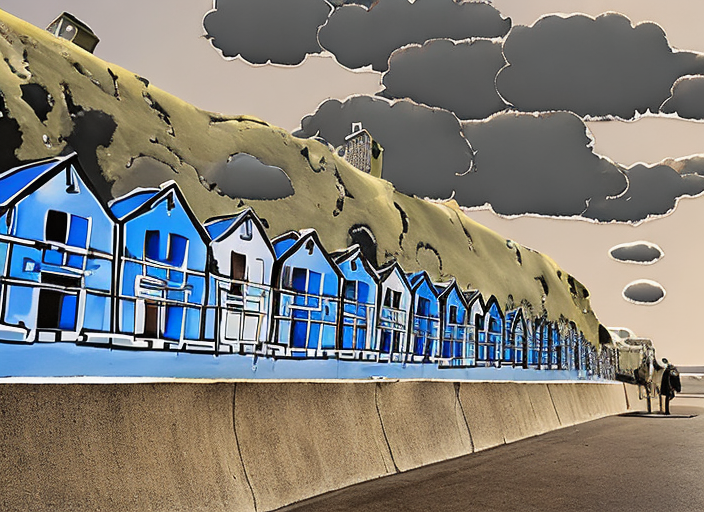}&
    \includegraphics[width=\linewidth]{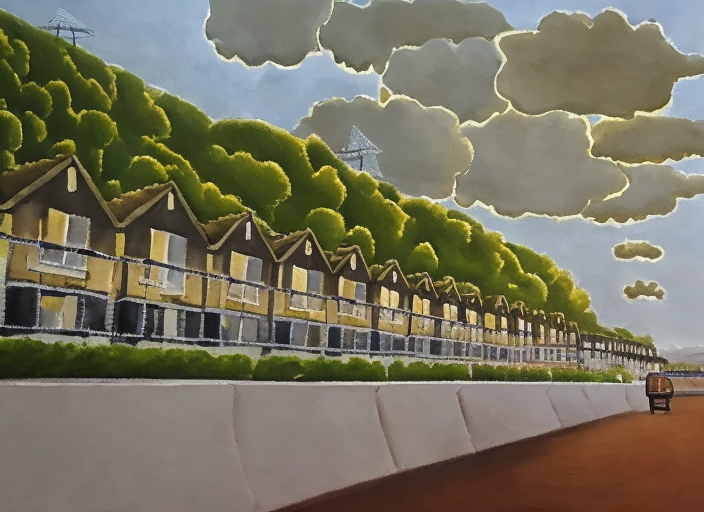}&
    \includegraphics[width=\linewidth]{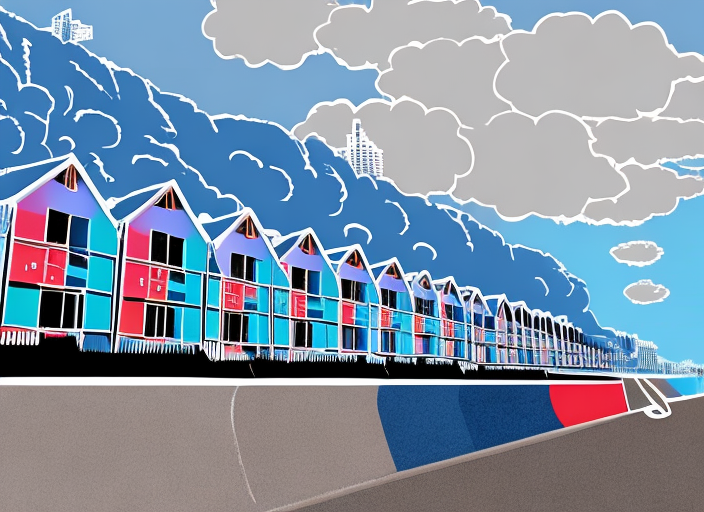}&
    \includegraphics[width=\linewidth]{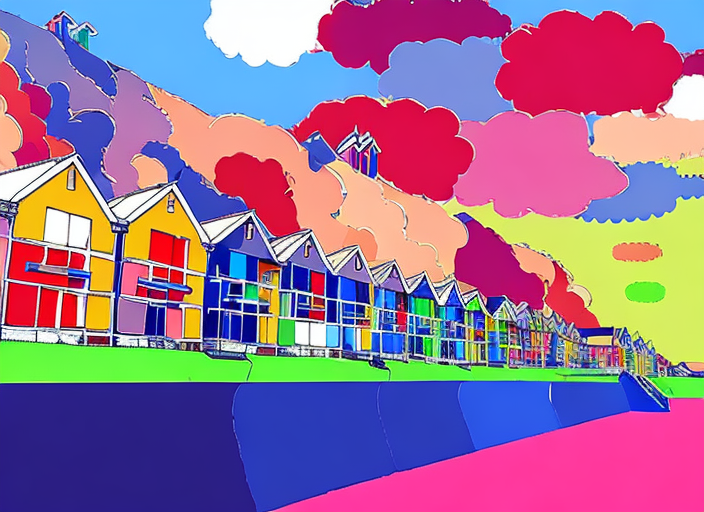}\\
    \includegraphics[width=\linewidth]{line-input-4}&
    \includegraphics[width=\linewidth]{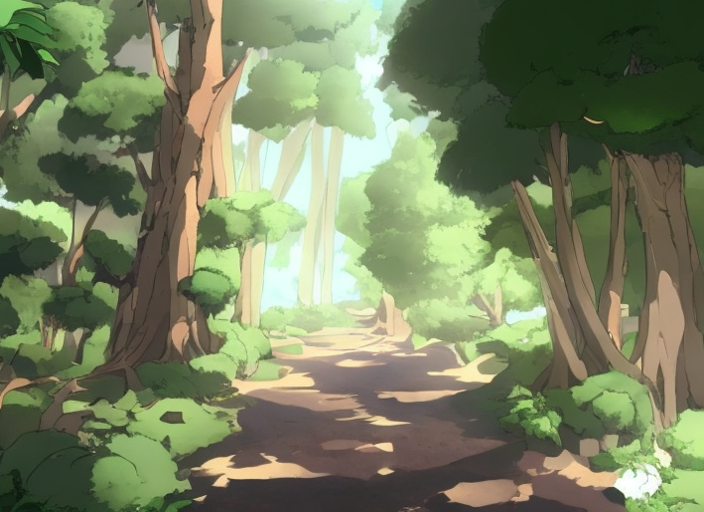}&
    \includegraphics[width=\linewidth]{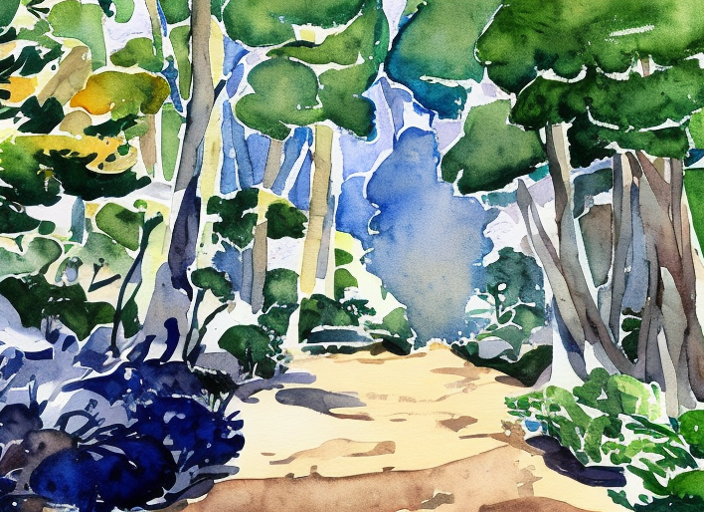}&
    \includegraphics[width=\linewidth]{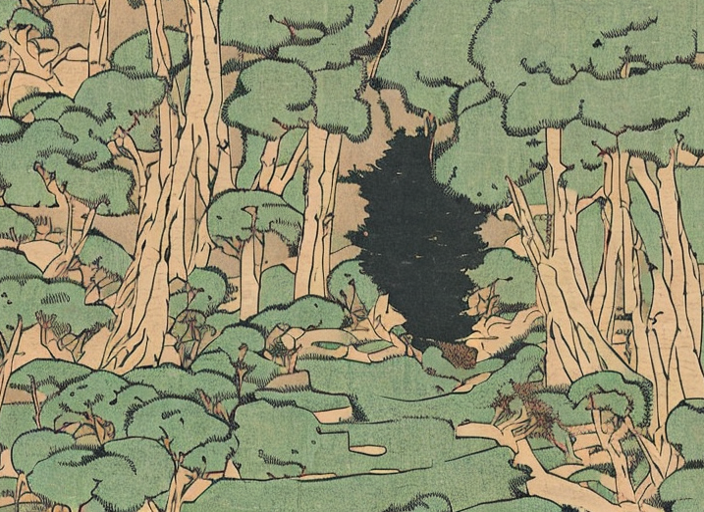}&
    \includegraphics[width=\linewidth]{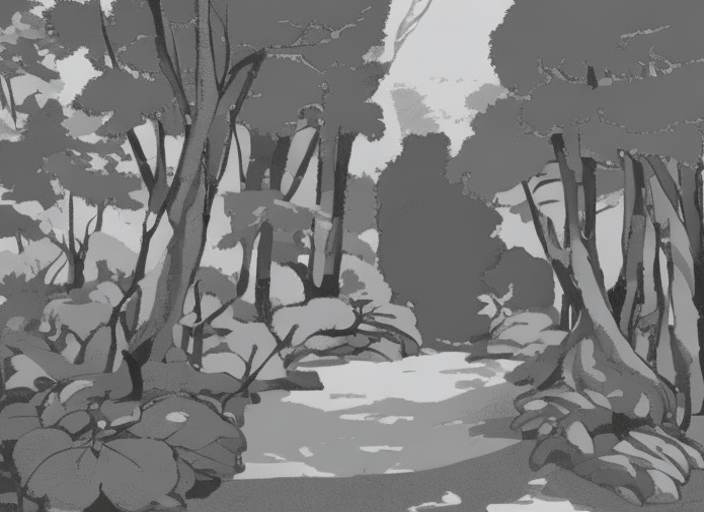}&
    \includegraphics[width=\linewidth]{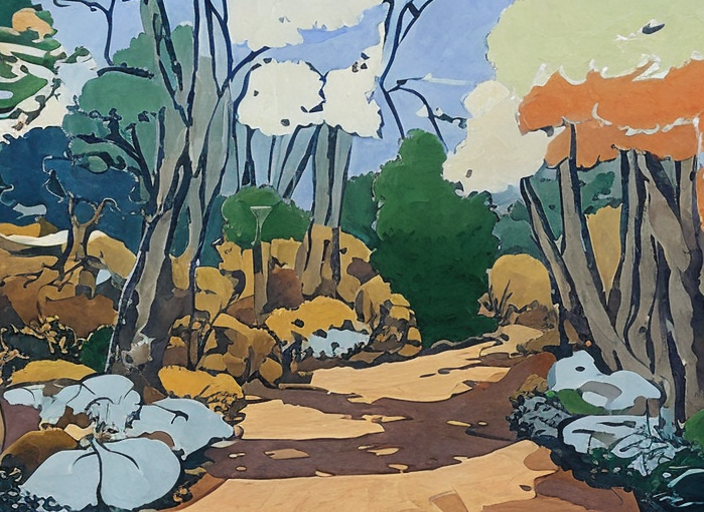}&
    \includegraphics[width=\linewidth]{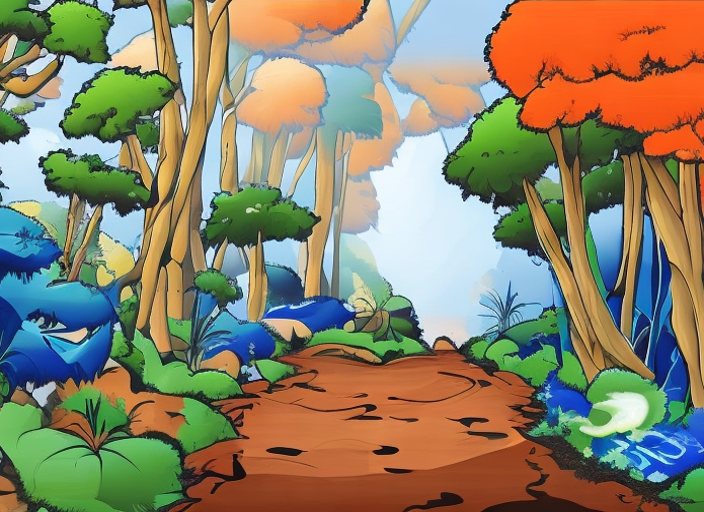}&
    \includegraphics[width=\linewidth]{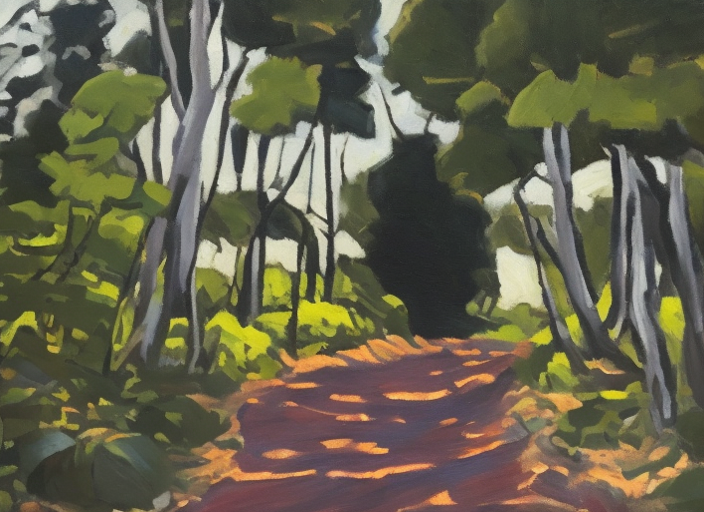}&
    \includegraphics[width=\linewidth]{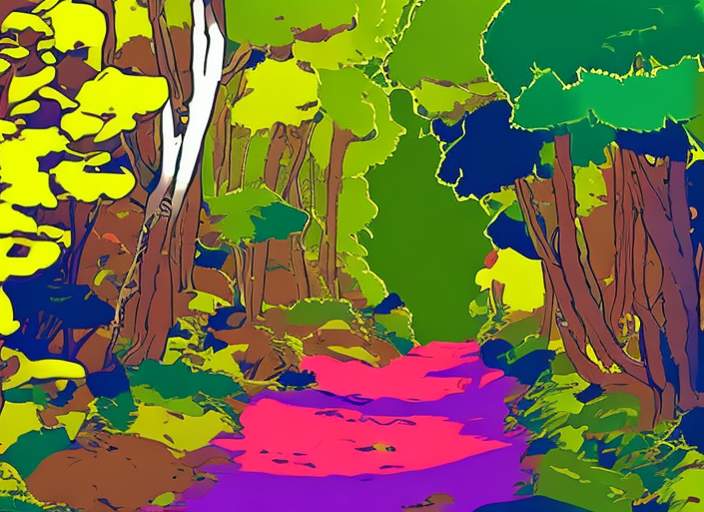}&
    \includegraphics[width=\linewidth]{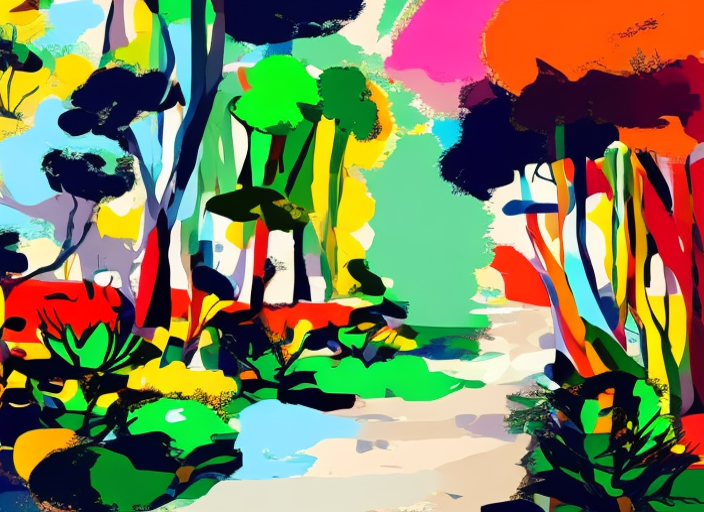}\\
    \end{tabular}
    }
    \vspace{-.1in}
    \caption{Synthesized images of line drawings with different text prompts by one-channel concatenation model.\label{fig:style}}
\end{figure*}

\subsection{Results}
We compare our fine-tuned models with PITI~\cite{wang2022pretraining}, which is a state-of-the-art diffusion model with reference-free scene sketch synthesis.
The PITI model was trained on the COCO-Stuff dataset~\cite{caesar2018coco} with HED~\cite{xie2015holistically} sketches and used the GLIDE~\cite{nichol2021glide} hierarchical generation scheme with two models: a base diffusion model at the resolution of $64 \times 64$ and a diffusion upsampling model for the resolution of $256\times256$.

\sisetup{table-parse-only,detect-weight=true,round-mode=places,round-precision=2}
\begin{table}[tbp]
  \caption{Reconstruction performance measured in LPIPS ($\downarrow$)~\cite{zhang2018unreasonable}. The scores were calculated over $300$ images of each GeoPose3K and LSUN-Church dataset under the size of $256\times256$.\label{tab:lpips}}
  \centering
  \resizebox{\columnwidth}{!}{%
  \begin{tabular}{lSS}
  \toprule
  {Method} & {GeoPose3K} & {LSUN-Church}\\
  \midrule
  {One-channel Concatenation} & 0.493 & 0.496\\
  {Three-channel Concatenation} & \bfseries \num{0.414} & \bfseries \num{0.416}\\
  {PITI$^{\dagger}$~\cite{wang2022pretraining}} & 0.691 & 0.447\\
  \bottomrule
  \multicolumn{3}{l}{$^{\dagger}$ \scriptsize The score for PITI was calculated over images generated from HED~\cite{xie2015holistically} edges as per pre-trained models.}
  \end{tabular}
}
\end{table}

\vspace{2mm}\noindent{\bf Quantitative results:}
Table~\ref{tab:lpips} summarizes the LPIPS score for the two datasets.
Overall, our three-channel concatenation model achieved the lowest LPIPS score for both test datasets.
In particular, PITI~\cite{wang2022pretraining} does not perform well on the GeoPose3K test set.

\vspace{2mm}\noindent{\bf Visual results:}
Subjectively, we also evaluate the visual quality of synthesized images for $50$ line images that were independently collected from the LAION dataset.
Since HED~\cite{xie2015holistically}, originally used as an edge detector in PITI, cannot detect edges of line images well, thus, we use DexiNed~\cite{poma2020dense} for the subjective evaluation.
Figure~\ref{fig:comparison} shows the visual comparison of reconstructed images for the GeoPose3K dataset, LSUN-Church, and line images.
Although the three-channel concatenation model provided higher objective evaluation performance, the one-channel concatenation model showed subjectively better aesthetic images.
When using DexiNed sketches, the PITI yielded a slight domain gap and produced less photorealistic images.
The styles were given as text prompts prefixed with ``a color photograph of''.
Our fine-tuned models show compelling results compared to the PITI, which has a domain gap for DexiNed sketches.

Since the one-channel concatenation model produced better aesthetic images, we further experimented with different text prompts such as ``an anime scene of'', ``a watercolor painting of'', ``an ukiyo-e art of'', ``a black and white photograph'' of, and so on to control the style of the output.
Synthesized images of line drawing images from Fig.~\ref{fig:comparison} with such text prompts are shown in Fig.~\ref{fig:style}.
Experiment results show that our fine-tuned models can generate images in accordance with given styles as text prompts.

\vspace{-4mm}
\section{Conclusion}
\vspace{-3mm}
In this paper, we show that by utilizing a standardized edge domain, challenging sketch-to-photo synthesis tasks can be tackled by easily fine-tuning pre-trained large-scale text-to-image models.
We carried out experiments and confirmed that line images (human-drawn sketches) are successfully converted to color images in different styles, where the target style can be intuitively controlled by text guidance.
As for future work, we shall further experiment with various kinds of sketches with various content, and investigate guidance models during sampling.

\begin{small}
\bibliographystyle{IEEEbib}
\bibliography{refs}
\end{small}

\end{document}